%% file: cvpr.tex
\documentclass[final]{cvpr}

\usepackage{times}
\usepackage{epsfig}
\usepackage{graphicx}
\usepackage{amsmath}
\usepackage{amssymb}
\usepackage{color}
\usepackage{gensymb}
\usepackage[ruled, vlined]{algorithm2e} % For algorithms
\usepackage[abs]{overpic}
\usepackage{url}
\usepackage{cite}
\usepackage{bm}
\newcommand{\para}[1]{\noindent\textbf{#1}.}
\newcommand{\paras}[1]{\noindent\textbf{#1}}
\usepackage{paralist}
\usepackage{float}
\usepackage{lipsum}

\usepackage[pagebackref=true,breaklinks=true,colorlinks,bookmarks=false]{hyperref}

\def\cvprPaperID{1824} % *** Enter the CVPR Paper ID here
\def\confYear{CVPR 2021}

\begin{document}

%%%%%%%%% TITLE
\title{Point2Skeleton: Learning Skeletal Representations from Point Clouds}

\author{Cheng Lin$^1$ \quad Changjian Li$^{2*}$ \quad  Yuan Liu$^1$ \quad Nenglun Chen$^1$ \quad Yi-King Choi$^1$ \quad Wenping Wang$^{1,3*}$ \\[0.3em]
$^1$The University of Hong Kong \quad $^2$University College London \quad $^3$Texas A\&M University 

}
\maketitle

\newcommand\blfootnote[1]{%
\begingroup
\renewcommand\thefootnote{}\footnote{#1}%
\addtocounter{footnote}{-1}%
\endgroup
}

\blfootnote{* corresponding authors}

%%%%%%%%% ABSTRACT
\begin{abstract}
We introduce Point2Skeleton, an unsupervised method to learn skeletal representations from point clouds. Existing skeletonization methods are limited to tubular shapes and the stringent requirement of watertight input, while our method aims to produce more generalized skeletal representations for complex structures and handle point clouds. Our key idea is to use the insights of the medial axis transform (MAT) to capture the intrinsic geometric and topological natures of the original input points. We first predict a set of skeletal points by learning a geometric transformation, and then analyze the connectivity of the skeletal points to form skeletal mesh structures. Extensive evaluations and comparisons show our method has superior performance and robustness. The learned skeletal representation will benefit several unsupervised tasks for point clouds, such as surface reconstruction and segmentation. 
\end{abstract}

%%%%%%%%% BODY TEXT
\section{Introduction}
\label{sec:introduction}
Generating skeleton-based representations to capture the underlying shape structures is a classic problem in computer vision and computer graphics. Skeletonization has been shown to benefit various tasks including shape recognition \cite{bai2009activeskel, trinh2011skeletonsearch}, 3D reconstruction \cite{wu2015deep, tang2019skeleton}, segmentation \cite{lin2020seg}, shape matching \cite{siddiqi1999shock, sundar2003skeletonbasedmatching}, pose estimation \cite{shotton2011realtimepose,newell2016stackedpose}, action recognition \cite{ke2017skel3dreg, si2018skeletonreg} and animation \cite{baran2007automaticanimation}. Extracting skeletons of 3D shapes using hand-crafted rules \cite{ma2003skeletonradius, au2008skeletonextraction, tagliasacchi2012meanskel} has been researched for decades.  With the recent advances in 3D vision with deep learning, predicting curve skeletons for 3D shapes using networks \cite{xu2019predictingskeleton} is beginning to be studied. 

In fact, the existing methods only target a specific category of shapes that can be abstracted appropriately by curve segments, i.e., shapes composed of tubular parts. Also, generating an internal representation, such as the skeleton, for a 3D shape heavily relies on watertight surface meshes to explicitly give inside/outside classification labels and precisely compute certain geometric functions \cite{chen2019learningimplicit, genova2019shapetemplate}. These restrictions critically limit the applicability of the existing skeletonization methods. Hence, it is imperative to develop an effective method for computing a generalized skeletal representation for an arbitrary 3D shape. Such a skeletonization method should be able to handle general input beyond the closed surface, such as point clouds with missing parts, in order to significantly extend the utility of skeleton in various computer vision and graphics tasks.

\begin{figure}[t]
    \centering      
    \vspace{-2mm}
  \begin{overpic}[width=\linewidth]{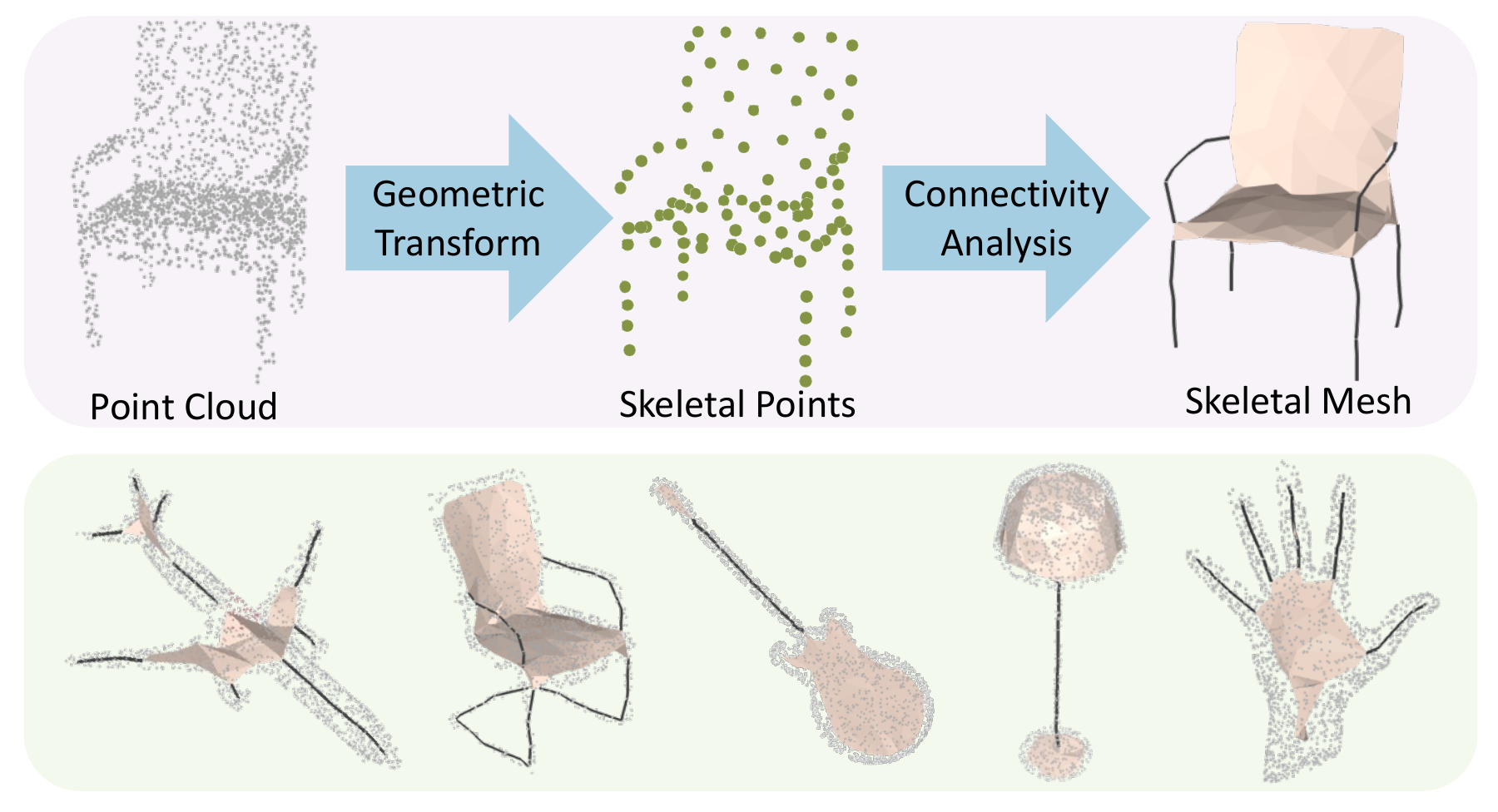}
    \end{overpic}
  \caption{We introduce an unsupervised method to learn skeletal meshes from point clouds. The skeletal meshes contain both 1D curve segments and 2D surface sheets which can represent underlying structures of various shapes. }
  \label{fig:teaser}
  \vspace{-3mm}
\end{figure} 

\begin{figure}[!htb]
\vspace{-2mm}
    \centering      
  \begin{overpic}[width=0.95\linewidth]{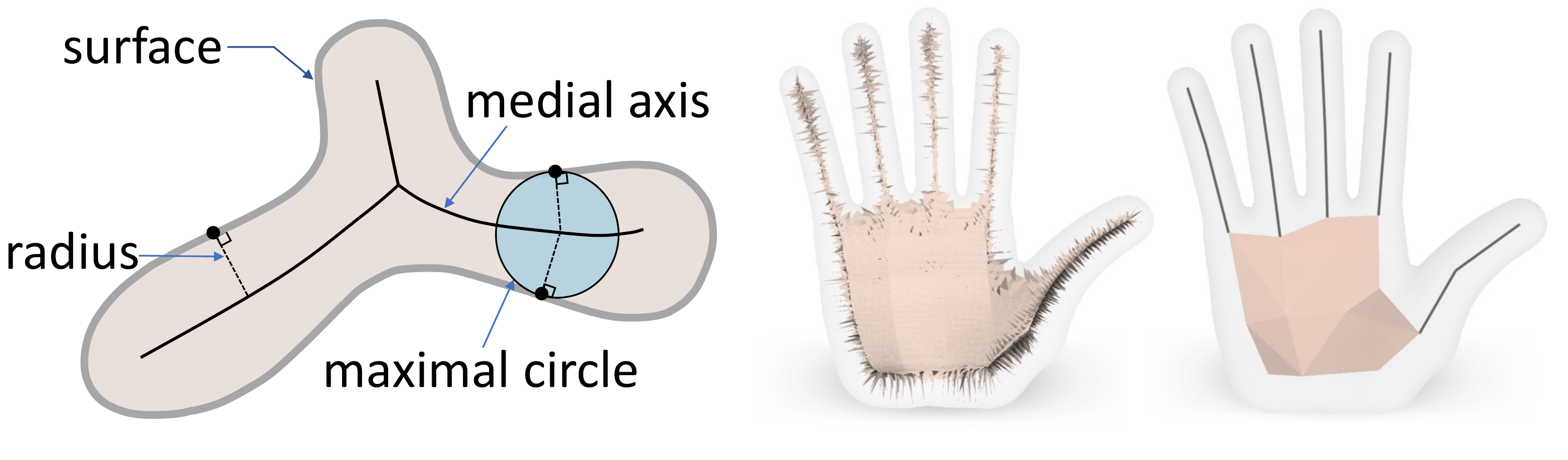}
    \put(55,-2){(a)}        
    \put(128,-2){(b)}
    \put(185,-2){(c)}
    \end{overpic}
  \caption{Illustration of the medial axis transformation (MAT) and the skeletal mesh. (a) The MAT of a 2D shape; (b) the original MAT of a 3D shape; (c) the skeletal mesh.}
  \label{fig:mat_intro}
\end{figure} 

We observe that the medial axis transform (MAT), one of the best known examples of skeletal representation, has a rigorous mathematical definition for arbitrary shapes, unlike the curve skeleton which is only empirically understood for tubular objects. Given a 3D shape, the MAT \cite{blum1967transformation} is defined as the set of points in the interior with more than one closest point on the boundary surface; the MAT encodes the shape to a lower-dimensional representation with the associated radius function. The examples are shown in Fig.~\ref{fig:mat_intro}. Despite its simple definition, the MAT is difficult to use in practice for the following two reasons: (1) the computation of the MAT is expensive, since it requires the input 3D shape to be defined by a closed boundary surface and a substantial amount of time for geometric processing; (2) the MAT is notoriously sensitive to surface noise, i.e., small perturbations to the shape surface lead to numerous insignificant branches \cite{amenta2001power, giesen2009SAT} (Fig.~\ref{fig:mat_intro}~(b)); such a noisy MAT does not clearly reflect the structures of the given shape.

These difficulties motivate us to resort to the formulation of the MAT and the representational power of the deep neural network to learn a generalized skeletal representation. We name this representation {\em skeletal mesh}, which is an important extension to the curve skeleton. As shown in Fig~\ref{fig:mat_intro}, the skeletal mesh follows a similar but not identical definition to the MAT; it circumvents the drawbacks of the MAT and has its own merits: (1) it is structurally meaningful and topologically informative, where the tubular parts can be properly abstracted by a few curves while the planar or bulky parts by interior surfaces; (2) the skeletal mesh is simpler and more compact than the standard MAT; it focuses on the fundamental geometry of a shape and can cope with point clouds, making it robust to surface noise and partially missing data. Therefore, given such a representation, the expressive capacity for the geometry and topology of a 3D shape is significantly enhanced.

In this paper, we propose {\em Point2Skeleton}, an unsupervised method for learning skeletal meshes from 3D point clouds. Our method consists of two main steps as shown in Fig~\ref{fig:teaser}. The first step is to predict the skeletal points by learning a geometric transformation. The second step is to connect the skeletal points to form a mesh structure; we adopt a graph structure, and analyze the edge connectivity by jointly leveraging the properties of skeletal mesh and the correlations learned by a graph auto-encoder (GAE). Our main contributions are:

\begin{compactitem}
    
    \item To our best knowledge, {\em Point2Skeleton} is the first unsupervised learning method for generalized point cloud skeletonization.
    
    \item We present novel unsupervised formulations for geometric learning of 3D point clouds, i.e., learning intrinsic geometric transformations and predicting connectivity for mesh generation. 
    
    \item We introduce a new representation, called {\em skeletal mesh}, which gives new insights into some unsupervised tasks for point clouds, such as surface reconstruction and segmentation.
    
\end{compactitem}

\begin{figure*} 
    \centering      
  \begin{overpic}[width=0.91\linewidth]{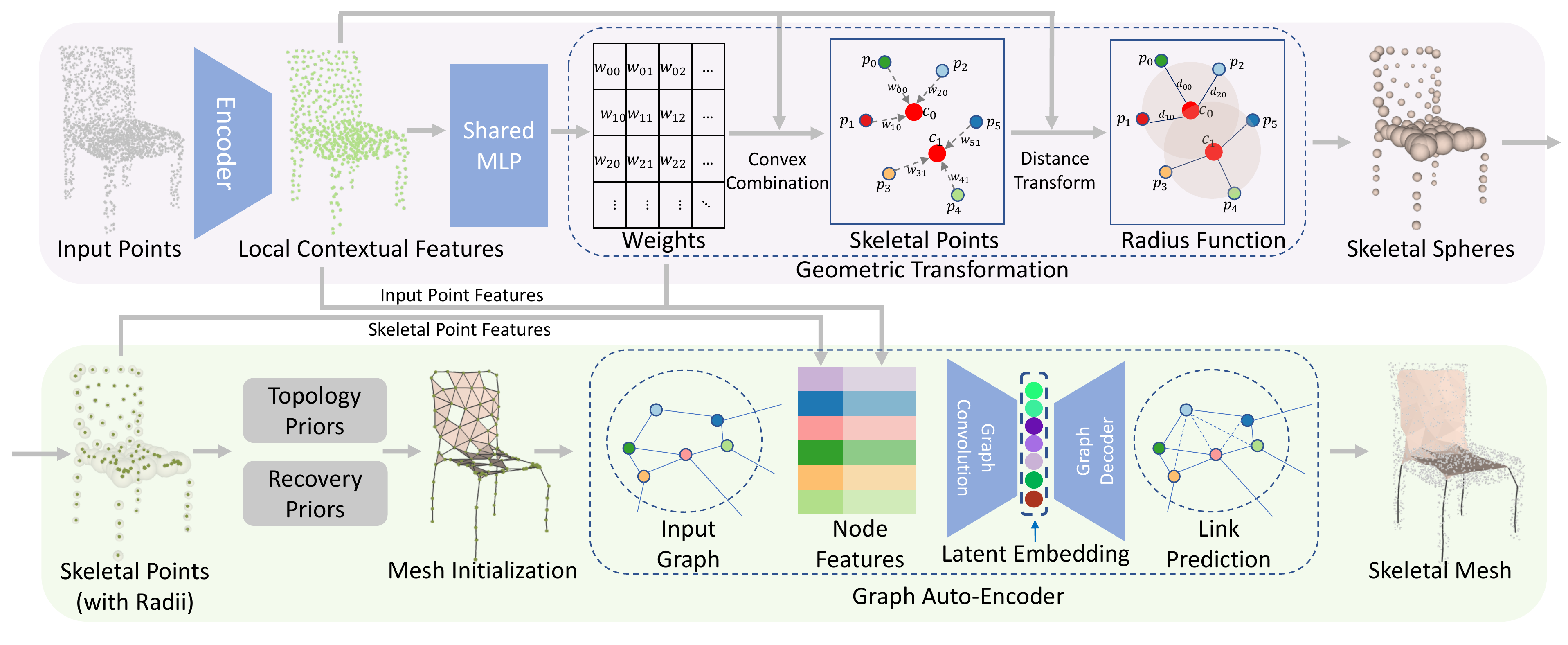}
    \end{overpic}
    \vspace{-2mm}
  \caption{Overview of our Point2Skeleton pipeline. Given a point cloud as input, first, we learn a geometric transformation via convex combinations to predict the skeletal points together with their radii. Second, we connect the skeletal points to form a mesh structure; we initialize a graph structure using two simple priors and formulate a link prediction problem using a graph auto-encoder to obtain a complete skeletal mesh. }
  \label{fig:pipeline}
  %\vspace{-4mm}
\end{figure*}

\section{Related Work}

\para{Skeletonization and medial axis transform} Skeleton-based shape representation has been extensively researched in computer vision and computer graphics. A widely-used form is the curve skeleton due to its simplicity. Traditional methods use hand-crafted rules to capture geometric properties for generating curve skeletons \cite{ma2003skeletonradius, au2008skeletonextraction, cao2010point, tagliasacchi2012meanskel, huang2013l1medial}. Recently, attempts have been made to use deep neural networks to predict curve skeletons \cite{xu2019predictingskeleton}. The curve skeletons are only empirically understood for tubular geometries, so they can only be applied to a limited class of shapes. 

Another form of skeletal representation is known as the medial axis transform (MAT) \cite{blum1967transformation}, which is a principled formulation that can encode arbitrary shapes. There are numerous methods for computing the MAT of 3D shapes, such as \cite{amenta1999voronoi}\cite{amenta2001power} and \cite{giesen2009SAT}. However, the MAT is notoriously sensitive to surface noise, which usually leads to numerous insignificant spikes. Therefore, some methods use simplification techniques to produce a clean and structurally simple MAT \cite{sun2015medial, li2015qmat, yan2018voxelcore}. These methods all need expensive geometric processing and require watertight input surfaces. There are also some rule-based methods for generating meso-skeletons \cite{tagliasacchi2012meanskel,wu2015deep} to approximate the MAT. We use deep learning techniques to efficiently predict the clean skeletons based on the insights of the MAT, and also show better performance than the other related methods.

\para{Learning transformations} 
Learning spatial transforms in the image domain is an important problem. Representative works on this problem include spatial transformer networks (STN) \cite{jaderberg2015spatial} and view synthesis by appearance flow \cite{zhou2016viewsythesis}, to name a few. However, learning geometric transforms in the 3D domain is still under-explored. Berkiten et al. \cite{berkiten2017learning} propose to learn a combination of geometric features to derive the transform of surface details. More relevantly, Yin et al. \cite{yin2018p2p} introduce a deep neural framework to learn geometric transforms between two domains of 3D points. Yang et al. \cite{yang2020p2mat} explicitly predict the medial axis from an input point cloud. These two methods are able to learn geometric transformations, but the learning is driven by the pre-computed ground-truth points rather than capturing the intrinsic geometric natures. Also, the learned point-to-point transformation neither contains structured surfaces nor gives topological connectivity.

\para{Inferring mesh structures} There has been growing interest in inferring mesh structures of 3D shapes using deep neural networks. Due to the difficulty of directly generating such compact structures, many approaches generate target meshes by deforming the pre-defined mesh templates \cite{litany2018deformable, wang2018pixel2mesh}. These methods lack the flexibility of handling complex geometries and typologies. Recently, learning implicit representations becomes popular. The common form is to first learn implicit volumetric functions to represent a 3D shape, such as signed-distance functions \cite{park2019deepsdf}, indicator functions \cite{mescheder2019occupancy} and structured implicit functions \cite{genova2019shapetemplate}, and then extract the surface. Some methods aim to directly predict mesh structures from scratch. Scan2Mesh \cite{dai2019scan2mesh} learns to explicitly generate vertices, edges and faces using graph neural networks. PointTriNet \cite{sharp2020pointtrinet} explicitly learns the triangulation of a given point set to form the mesh.

Most of these methods rely on ground-truth surfaces for reliable supervision; none of them can be used to predict skeleton representations that contain curves. 
In contrast, our method is directly trained on point clouds, aiming to effectively generate skeletal meshes in an unsupervised manner.

\section{Skeletal Mesh}
\label{sec:skl_mesh}

\begin{figure}[!t]
  \vspace{-6mm}
    \centering      
  \begin{overpic}[width=0.9\linewidth]{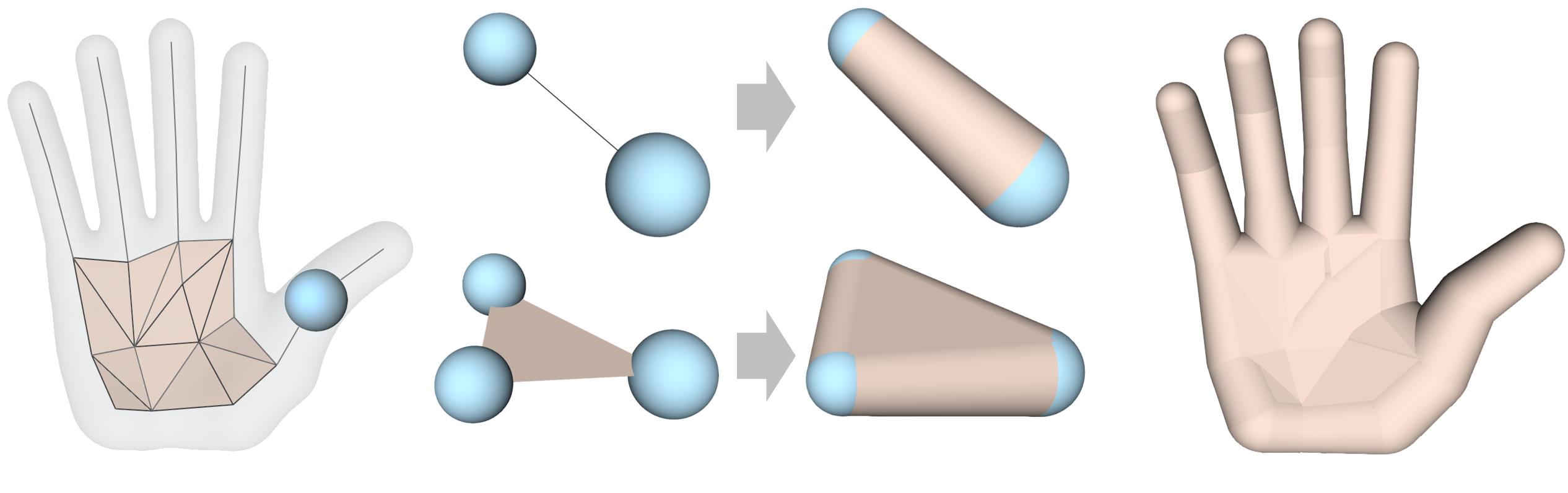}
    \put(16,-1){ (a)}
    \put(100,-1){ (b)}
    \put(175,-1){ (c)}
    \end{overpic}
  \caption{Properties of skeletal mesh. (a) A skeletal mesh structure of a shape; (b) interpolation of the skeleton spheres of an edge (top) and a triangle face (bottom); (c) the interpolation of all the spheres on the skeletal mesh reconstructs the original shape.}
  \label{fig:prelinimary}
  \vspace{-4mm}
\end{figure}

The skeletal mesh of a given 3D shape is a discrete 2D non-manifold defined by a collection of skeletal points, edges and faces that form the underlying structure of the 3D shape, as shown in Fig.~\ref{fig:prelinimary}~(a). A {\em skeletal sphere} is required to be maximally inscribed in the 3D shape, denoted as $s=(c,r(c)) \in \mathbb{R}^4$, where $c \in \mathbb{R}^3$ is the center of the sphere (i.e., the {\em skeletal point}), and $r(c): \mathbb{R}^3 \rightarrow \mathbb{R}$ is the associated {\em radius}. We use $e_{ij}=(c_i, c_j)$ to represent an edge of the skeletal mesh connecting two skeletal points $c_i$ and $c_j$, and use $f_{ijk}=(c_i, c_j, c_k)$ to represent a triangle face. 

The skeletal mesh is a discrete representation while the MAT is continuously defined. The discrete nature of the skeletal mesh allows us to develop a learning-based approach to robustly compute the skeletal representation, such that it inherits the favorable properties of the MAT for shape representation while not suffering from the instability to boundary noise \cite{amenta2001power, giesen2009SAT}. Several good properties of the skeletal mesh are elaborated as follows, which makes it a useful representation for shape analysis (see Sec.~\ref{sec:application}).

\para{Recoverability} The skeletal mesh can be considered as a complete shape descriptor, which means it can reconstruct the shape of the original domain. Fig.~\ref{fig:prelinimary}~(b)(c) shows how to reconstruct the original shape by the interpolation of all the skeletal spheres based on the mesh structure. 

\para{Abstraction} The skeletal mesh captures the fundamental geometry of a 3D shape and extracts its global topology; the tubular parts are abstracted by simple 1D curve segments and the planar or bulky parts by 2D surface triangles. 

\para{Structure awareness} The 1D curve segments and 2D surface sheets as well as the non-manifold branches on the skeletal mesh give a structural differentiation of a shape.

\para{Volume-based closure} The interpolation of the skeletal spheres gives solid cone-like or slab-like primitives (see Fig.~\ref{fig:prelinimary}~(b)); then a local geometry is represented by volumetric parts, which provides better integrity of shape context. The interpolation also forms a closed watertight surface.

\section{Method}
Our method is composed of two modules. As shown in Fig.~\ref{fig:pipeline}, given a point cloud as input, the first module is to predict a set of skeletal points by learning a geometric transformation (Sec.~\ref{sec:MAT_module}). The second module is to connect the skeletal points to form a skeletal mesh, where we resort to a graph structure to analyze the edge connectivity (Sec.~\ref{sec:link_pred_module}).

\subsection{Skeletal Point Prediction}
\label{sec:MAT_module}

Given $K$ input points $\{p_i\}$ with 3D coordinates, represented as $\bm{P} \in \mathbb{R}^{K\times 3}$, our goal is to predict $N$ skeletal spheres $\{s_i=(c_i, r(c_i))\}$, represented as $\bm{S} \in \mathbb{R}^{N\times 4}$, consisting of the coordinates of the centers $\bm{C} \in \mathbb{R}^{N\times 3}$ and their radii $\bm{R} \in \mathbb{R}^{N\times 1}$. To start with, we use PointNet++ \cite{qi2017pointnet++} as the encoder to obtain the sampled input points $\bm{P'} \in \mathbb{R}^{K'\times 3}$ and their contextual features $\bm{F} \in \mathbb{R}^{K' \times D}$, where  $K'\textless K$ and $D$ is the dimension of the feature.

\para{Convex combination} We observe a skeletal point can be considered as a local center of a set of surface points. Consequently, instead of directly predicting the coordinates of the skeletal points, we use the convex combination of input points to generate the skeletal points. To this end, we predict the weights $\bm{\mathcal{W}} \in \mathbb{R}^{K' \times N}$ of the sampled input points $\bm{P'}$ in the convex combination
and all the skeletal points $\bm{C}$ are derived by:
\vspace{-2mm}
\begin{equation}
    \bm{C} = \bm{\mathcal{W}}^T \bm{P'} \quad s.t.\ \   j\!=\!1,...,N \quad \sum_{i=1}^{K'} \bm{\mathcal{W}}(i,j)=1.
\label{eq:cvx_center}
\vspace{-2mm}
\end{equation}

Now we explain how to estimate the radius of each skeletal sphere. First, the closest distance from an input point $p$ to all the skeleton points $\{c_i\}$ is defined as follows:
\begin{equation}
    d(p, \{c_i\}) =  \min_{c \in \{c_i\}} \left\| p-c \right\|_2.
    \label{eq:r_dis}
\end{equation}

The distances for all $K'$ sampled input points ${p_i}$ computed by Eq.~\ref{eq:r_dis} are summarized in a vector $\bm{D} \in \mathbb{R}^{K' \times 1}$. Now we express the radii of all the skeletal points as the linear combinations of their closest distances of all the sampled points, i.e., $\bm{R}= \bm{\mathcal{W}}^T\bm{D}$. The rationale of using the same weights $\bm{\mathcal{W}}$ as Eq.~\ref{eq:cvx_center} is based on the fact that the predicted weights for a skeletal point $c$ are large only for the input points close to $c$, but diminish to zero for those far from $c$. The reader can refer to the supplementary material for a detailed argument.

The advantages of using the weights of convex combination to predict the skeletal points lie in: (1) the generated points tend to be located inside the local shape without using explicit inside/outside labels given by a watertight boundary surface; (2) the computed combinational weights $\bm{\mathcal{W}}$ can be considered as linear filters which are effective for denoising when computing the radius. 

Similar to  \cite{chen2020unsupervised}, we use a shared multi-layer-perceptron (MLP) followed by a softmax layer to produce the weights $\bm{\mathcal{W}}$. By capturing the properties of the skeletal mesh, we are able to design a set of loss functions to train the network to obtain the combinational weights $\bm{\mathcal{W}}$.

Based on the recoverability of the skeletal mesh, we first introduce two loss functions to measure the reconstruction error from two different perspectives.

\para{Sampling loss} We sample points on the surface of each skeletal sphere and measure the Chamfer Distance (CD) between the sampled points $\{t_i\}$ and the input points $\{p_i\}$:
\begin{equation}
    L_{s}=\sum_{p \in \{p_i\}} \min_{t \in \{t_i\} } \left\|p - t \right\|_2 +\sum_{t\in \{t_i\}} \min_{p \in \{p_i\}} \left\|t - p \right\|_2.
\end{equation}
Here we adopt a uniform sampling strategy. For a skeletal sphere $(c, r(c))$, a surface point $t$ can be obtained by $t=c+r(c)v$ given a unit direction vector $v$; we sample 8 points for each skeletal sphere by giving 8 unit vectors $V=\{(\pm \eta, \pm \eta, \pm \eta), 3\eta^2=1\}$.

\para{Point-to-sphere loss} In addition to sampling, we introduce a loss function that measures the reconstruction error by explicitly optimizing the coordinates of the skeletal points and their radii:
\begin{equation}
    L_{p}\!\!=\!\!\!\!\sum_{p\in \{p_i\}}\!\!(\!\!\!\min_{c \in \{c_i\}}\!\left\|p\!-\!c\right\|_2\!-r(c^{min}_p)\!)+\!\!\!\!\sum_{c\in \{c_i\}}\!\!(\!\!\!\min_{p \in \{p_i\}}\! \left\|c\! -\! p \right\|_2 \!- r(c)\!),
\end{equation} 
where $\{c_i\}$ represents the predicted skeletal points, $\{p_i\}$ the input points, $r(c)$ the radius of a skeletal point $c$, and $c^{min}_p$ the closest skeletal point to the input point $p$. The first term constrains each input point to be located on the surface of its closest skeletal sphere, while the second term encourages each skeletal sphere to touch its closest input point. 

These two complementary losses optimize the consistency between the skeleton and the input point cloud from different aspects, leading to more reliable predictions (see the evaluations in Sec.~\ref{sec:ablation}).

\para{Radius regularizer} An inscribed sphere of a 3D shape is sensitive to surface noise; that is, a sphere can be stuck in the space among several noisy points thus resulting in a tiny radius. To focus on the fundamental geometry for better abstraction and avoid instability, therefore, we also include a radius regularization loss to encourage larger radii:
\begin{equation}
    L_{r}=-\sum_{c\in \{c_i\}} r(c).
\end{equation}
\vspace{-2mm}

Instead of directly being predicted by the network, the radii are computed using a linear combination of the distances in Eq.~\ref{eq:r_dis} with the weights $\bm{\mathcal{W}}$. Thus, the radius values are bounded to a certain range, which prevents them from growing too large and giving a very small value of $L_r$ to destroy the optimization. The total loss function for predicting skeletal spheres is a weighted combination of these three terms:
\begin{equation}
    L_{skel}=L_s+ \lambda_1 L_p+\lambda_2 L_r.
\label{loss_skel}
\end{equation}

\subsection{Skeletal Mesh Generation}
\label{sec:link_pred_module}
In this section, we aim to connect the predicted skeletal points to form skeletal meshes. Our mesh generation process is built on the edge connectivity of the skeletal points, which can encode both curve segments and face triangles. We thus resort to a graph representation for this problem.

\para{Graph initialization}
Based on the properties of the skeletal mesh, we first initialize a set of reliable links using two simple priors:

\begin{compactitem}
    \item \paras{Topology prior:} A node has a link to its closest node (marked as {\em known existing links}) but no links to its $k$-farthest nodes ({\em known absent links}). 
    
    \item \paras{Recovery prior:} Given an input point $p$ with its two closest skeletal points $c_1, c_2$, there will be a link connecting $(c_1, c_2)$ ({\em known existing links}).
\end{compactitem}

The topology prior assumes the skeletal mesh is relatively regular. The recovery prior ensures each input point can be properly reconstructed by the skeletal mesh. Based on these reliable priors, we obtain an initialized graph with {\em known existing} and {\em known absent} links, and we mark the other links as {\em unknown}.

\begin{figure}[!t]
\vspace{-2mm}
    \centering      
  \begin{overpic}[width=\linewidth]{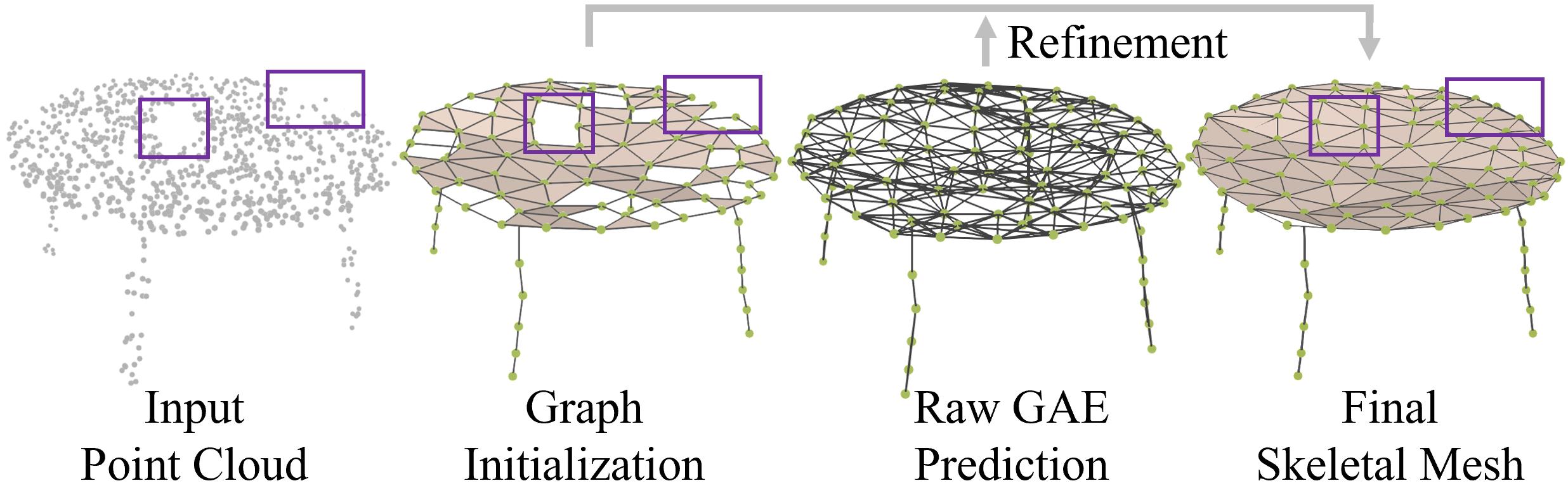}
    \end{overpic}
  \caption{Skeletal mesh generation. Given the initialized graph, the GAE predicts potential links by measuring the correlations of the skeletal points in the latent embedding, based on which we refine the initialized graph to derive the final skeletal mesh.}
  \label{fig:link_pred}
  \vspace{-2mm}
\end{figure}

\para{Link prediction by GAE} 
The initialized graph is usually incomplete. Therefore, we formulate a link prediction problem based on the graph auto-encoder (GAE) \cite{kipf2016variational}, which is an unsupervised formulation, to further predict the missing links by analyzing the correlations of the skeletal points in the latent space. As shown in Fig.~\ref{fig:link_pred}, the GAE predicts additional links for the initialized graph, enabling our method to handle missing regions and noise of the input points. 

The input to the encoder is the initialized graph $\mathcal{G}=(\mathcal{N}, \mathcal{E})$ with $N=|\mathcal{N}|$ nodes representing $N$ skeletal points and $\mathcal{E}$ the edge connectivity. The graph $\mathcal{G}$ is undirected and unweighted, represented by an adjacency matrix $\bm{A}\in \{0, 1\}^{N\times N}$. To effectively leverage the geometric correlations between the skeleton and the input surface points, the node features are jointly characterized by the coordinates $\bm{C}$ of the skeleton points, their radii $\bm{R}$ and the contextual features $\bm{F}$ of the input, which is denoted as $[\bm{C}, \bm{R}, \bm{\mathcal{W}}^T\bm{F}]$, where $[\cdot,\cdot, \cdot]$ represents the concatenation on the feature dimension and $\bm{\mathcal{W}}^T\bm{F}$ combines the contextual features to the skeletal points using the predicted weights $\bm{\mathcal{W}}$.

The encoder is a series of graph convolutional network (GCN) layers with residual blocks \cite{he2016deep} between the consecutive layers. We use an inner product decoder to produce the reconstructed adjacency matrix $\bm{\hat{A}}$. Please refer to the supplementary material to find more details.

\para{Learning} We define a reconstruction loss that captures the similarity of the reconstructed $\bm{\hat{A}}$ and the initialized $\bm{A}$:
\begin{equation}
    L_{link}\!\!=\!\!\text{mean}(\mathcal{M} \odot ( -\xi \bm{A}\text{log}(\sigma(\bm{\hat{A}}))-(1\!-\!\bm{A}) \text{log}(\sigma(1\!-\!\bm{\hat{A}})))).
\end{equation}

This is referred to as a Masked Balanced Cross-Entropy (MBCE) loss \cite{tran2018pred_ae}. Here $\xi$ represents the balanced weight, which is the ratio of the amount of {\em absent} links to the {\em existing} links; $\mathcal{M}\in \mathbb{R^{N\times N}}$ is a binary mask indicating whether a link between two nodes is {\em known} or {\em unknown} (as marked in the graph initialization priors), by which we only consider the {\em known} links in the back-propagation; $\odot$ is the element-wise product and $\sigma(\cdot)$ represents the Sigmoid function. 

The capability of the GAE to predict missing links is attributed to the encoded meaningful latent embeddings that can reconstruct the graph. The decoder establishes the links by measuring the correlations of the latent features; thus the nodes with unknown connectivity will form links if they exhibit strong correlations in the latent space.

\para{Mesh generation} Once the training is finished, we do not directly use the raw output of GAE as the final skeletal mesh, since it will lead to many redundant triangles and unclear structures. Instead, as shown in Fig.~\ref{fig:link_pred}, we use the GAE prediction to refine the initial graph, which mainly contains two steps. The first is {\em hole filling}. We extract all the triangle faces on the initial graph, and then detect the small polygonal loops that do not form face triangles on the remaining initial graph. A polygonal loop hole will be filled if the links predicted by the GAE show the hole does not exist. To explicitly consider the geometric recoverability for more reliable mesh generation, we also fill a hole if there are enough points on its corresponding area of the input points. The second is {\em boundary refinement}. We extract the skeletal points on the boundary and connect two points if the GAE prediction shows they are connected. With the predicted mesh, we can re-compute the closest distances based on Eq.~\ref{eq:r_dis} and then obtain a more accurate radius estimation.

\begin{figure*}[!htb]
\centering      
  \begin{overpic}[width=\linewidth]{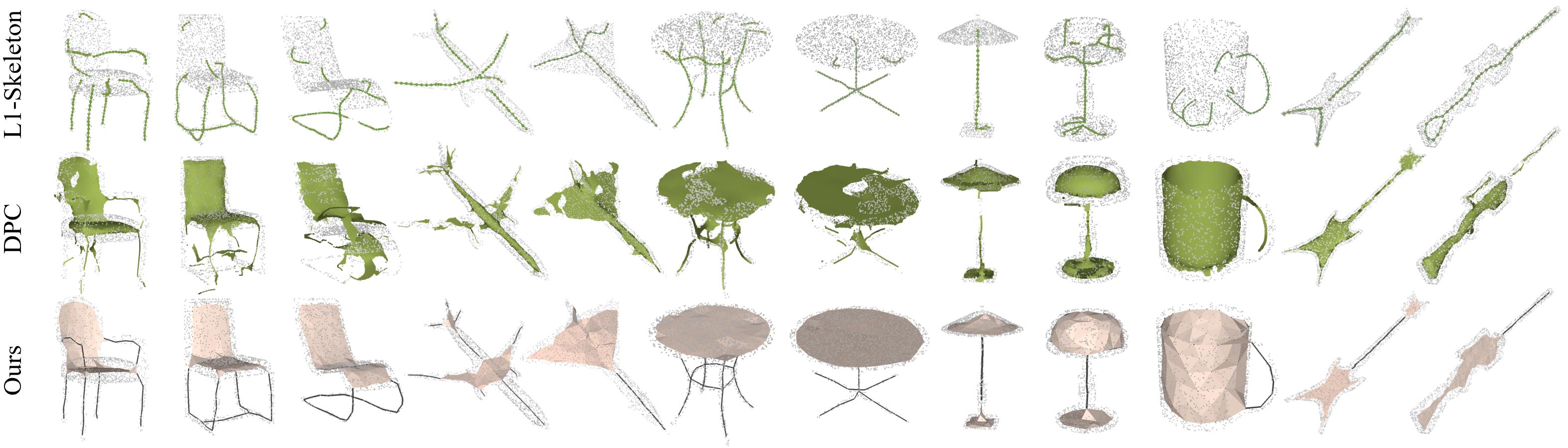}
    \end{overpic}
  \caption{Qualitative comparison with the competitive point cloud skeletonization methods, i.e., $L_1$-medial skeleton \cite{huang2013l1medial} and deep point consolidation (DPC) \cite{wu2015deep}.}
  \label{fig:cmp}
  \vspace{-4mm}
\end{figure*} 

\section{Experimental Results}

\para{Dataset} We collect 7088 shapes from 8 categories of ShapeNet  \cite{chang2015shapenet}, and use a virtual scanner \cite{Wangocnn2017} to generate the point clouds (2000 points are randomly sampled for each shape). We use 5/6 of the data from each category for training and the other 1/6 for testing. The network is trained on all the shape categories jointly. 

\para{Implementation details} We use a PointNet++ \cite{qi2017pointnet++} encoder of 4 set abstraction levels to obtain the contextual features; then the features are processed by 5 shared MLPs to predict $N\!\!=$100 skeletal points for each shape. For the GAE, we use a 12 layers GCN with residual blocks. More details and evaluations are given in the supplementary material.

For the training process, the skeletal point prediction network is first pre-trained only by the CD loss between the skeletal points and the input points for 20 epochs. This is to make the skeletal points evenly distributed around the shape, leading to a better initialization. Then, the network is trained using $L_{skel}$ (with $\lambda_1=0.3$ and $\lambda_2=0.4$) for 30 epochs. After that, we freeze the skeletal point prediction network and train the GAE for 30 epochs. The learning rates for the two networks are $1e^{-3}$ and $5e^{-4}$ respectively and the batch size is 4.

\subsection{Results and Comparisons} 

\para{Evaluation metrics} We evaluate our method and the competitive skeletonization methods from two aspects. The first is the reconstruction quality for the original shape. We measure the difference between the shapes reconstructed from the skeletons and the ground truth shapes using the Chamfer distance (CD) and Hausdorff distance (HD), which are denoted as {\em CD-Recon} and {\em HD-Recon}, respectively.

Note that only the reconstruction accuracy alone is not enough to completely reflect the quality of the skeletonization. For example, a poorly designed algorithm that directly outputs the input points will yield small reconstruction error, but do not constitute a reasonable skeleton. Unfortunately, there are no existing metrics to evaluate if a skeletonization is reasonable, since the skeleton is a form of shape abstraction involving higher-level perceptions. Nevertheless, we observe a manually simplified MAT, where the unstable spikes are removed, not only shares some common characteristics with different forms of skeletons, but also has good geometric accuracy. A simplified MAT contains both curve-like and surface-like geometries; thus it can be used to evaluate the methods that are based on curves or beyond the curves.

We repair the ground truth meshes of the ShapeNet, convert them to watertight meshes \cite{huang2018manifold} and compute the strictly defined MATs. We manually simplify the MATs using the handcrafted methods \cite{li2015qmat} by removing the spikes to make them visually simple and clean (see the supplementary material for more details). Again, we use the CD and HD to measure the difference between the output skeletal representations and the simplified MAT, denoted as {\em CD-MAT} and {\em HD-MAT}. All these distances are computed by randomly sampled points from the respective geometries.

\para{Comparisons} Since learning-based skeletonization is barely studied, we evaluate our method with comparisons to closely relevant approaches that are not learning-based, i.e., $L_1$-medial skeleton \cite{huang2013l1medial} and deep point consolidation (DPC) \cite{wu2015deep}. We use the code released by the authors for comparison. Similar to ours, these methods are designed for the skeletonization of an arbitrary 3D shape given as a point cloud.  Fig.~\ref{fig:cmp} shows the qualitative comparisons with these methods. The $L_1$-medial skeleton \cite{huang2013l1medial} can produce structured representations that only contain 1D curves, thus resulting in large errors when used to abstract non-tubular shapes. The DPC \cite{wu2015deep} method can generate both surface-like and curve-like skeletons, but the representations are unstructured points that lack topological constraints, leading to inconsistency for thin structures. In contrast, our method can generate more compact and structurally meaningful skeletal representations for various geometries. 

As shown in Table~\ref{tab:cmp}, we quantitatively compare our method with these two methods using the four metrics aforementioned. The results show that our method not only more accurately encodes the information from the original input, but also produces more reasonable skeletonization results that are geometrically meaningful.

\begin{table*}[]
\centering
\scalebox{0.9}{
\begin{tabular}{c|ccc|ccc||ccc|ccc}
\hline
 & \multicolumn{3}{c|}{CD-Recon} &  \multicolumn{3}{c||}{HD-Recon} & \multicolumn{3}{c|}{CD-MAT}  & \multicolumn{3}{c}{HD-MAT} \\\hline
         & L1       & DPC             & Ours            & L1       & DPC             & Ours            & L1     & DPC    & Ours            & L1     & DPC    & Ours            \\\hline
Airplane & 0.0378 & \textbf{0.0348} & 0.0363          & 0.2216 & 0.1436          & \textbf{0.1266} & 0.0793          & 0.1307 & \textbf{0.0611} & 0.2384          & 0.2580 & \textbf{0.1721} \\
Chair    & 0.1126 & 0.0769          & \textbf{0.0441} & 0.4810 & 0.2478          & \textbf{0.1618} & 0.1885          & 0.2286 & \textbf{0.0974} & 0.4991          & 0.3707 & \textbf{0.2151} \\
Table    & 0.1041 & 0.0853          & \textbf{0.0424} & 0.3453 & 0.2584          & \textbf{0.1745} & 0.1541          & 0.2683 & \textbf{0.0876} & 0.3583          & 0.3690 & \textbf{0.2085} \\
Lamp     & 0.1542 & 0.0712          & \textbf{0.0335} & 0.3956 & 0.1850          & \textbf{0.1382} & 0.1870          & 0.1751 & \textbf{0.0884} & 0.4089          & 0.2627 & \textbf{0.2003} \\
Guitar   & 0.0655 & 0.0212          & \textbf{0.0179} & 0.2180 & \textbf{0.0589} & 0.0625          & 0.0817          & 0.0672 & \textbf{0.0536} & 0.2262          & 0.1226 & \textbf{0.1216} \\
Earphone & 0.0437 & 0.0573          & \textbf{0.0399} & 0.1908 & 0.2059          & \textbf{0.1125} & \textbf{0.0607} & 0.2216 & 0.1638          & \textbf{0.1732} & 0.3403 & 0.2130          \\
Mug      & 0.2864 & 0.1280          & \textbf{0.0417} & 0.9142 & 0.3510          & \textbf{0.1419} & 0.5316          & 0.4600 & \textbf{0.1179} & 0.9057          & 0.4308 & \textbf{0.2158} \\
Rifle    & 0.0260 & 0.0215          & \textbf{0.0213} & 0.1078 & \textbf{0.0702} & 0.0767          & 0.0494          & 0.0427 & \textbf{0.0356} & 0.1234          & 0.1050 & \textbf{0.0957} \\
Average  & 0.1038 & 0.0668          & \textbf{0.0372} & 0.3593 & 0.2049          & \textbf{0.1424} & 0.1665          & 0.2026 & \textbf{0.0828} & 0.3667          & 0.3047 & \textbf{0.1898}\\\hline
\end{tabular}
}
\caption{Quantitative comparison with the competitive point cloud skeletonization methods. }
\label{tab:cmp}
\end{table*}

\subsection{Discussions}
\label{sec:ablation}
We conduct a series of ablation studies to verify the various settings in our framework, and also some experiments to further explore the properties of the proposed method. 

\begin{figure}[!tb]
    \centering      
  \begin{overpic}[width=0.95\linewidth]{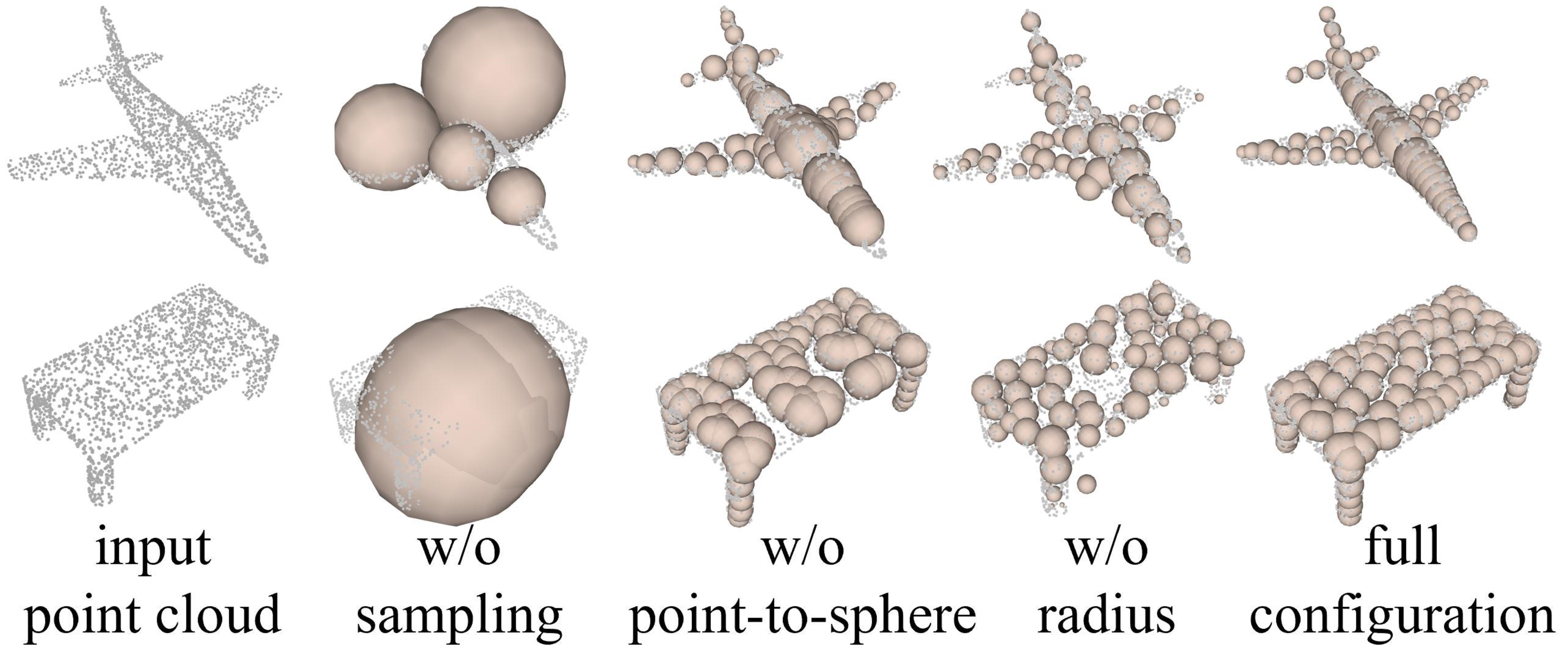}
    \end{overpic}
  \caption{Qualitative ablation study on different configurations of the geometric constraints for skeletal point prediction.}
  \label{fig:abl_skelpoints}
  \vspace{-4mm}
\end{figure}

\para{Skeletal point prediction} We evaluate the effect of the geometric constraints in the loss function (Eq.~\ref{loss_skel}) for skeletal point prediction. We alternatively remove each constraint and analyze the quality of the predicted skeletal points. The qualitative and quantitative results are shown in Fig.~\ref{fig:abl_skelpoints} and Table~\ref{tab:abl_skelpoints}, respectively. Under the effect of the radius regularizer that encourages larger radii, only the point-to-sphere loss struggles to constrain the skeletal spheres inside the shape without using sampling loss. The removal of the point-to-sphere loss results in inaccurate radii and uneven distribution of spheres, while the removal of the radius regularizer leads to spheres that are scattered instead of being maximally inscribed. Using the full configuration better captures the inner structures and achieves higher accuracy.

\begin{table}[!tb]
\vspace{-4mm}
\resizebox{\columnwidth}{!}
{\begin{tabular}{c|cccc}\hline
                  & CD-Recon        & HD-Recon        & CD-MAT          & HD-MAT          \\\hline
w/o sampling($L_s$)         & 0.3032               & 0.4886               & 0.7319               & 0.7848               \\
w/o point2sphere($L_p$)     & 0.0535               & 0.1619               & 0.1102               & 0.2236               \\
w/o radius($L_r$)           & 0.0634               & 0.2206               & 0.1276               & 0.2517               \\
full configuration   & \textbf{0.0525}      & \textbf{0.1592}      & \textbf{0.1060}      & \textbf{0.2079}      \\\hline
\end{tabular}}
\caption{Quantitative ablation study using different configurations of the geometric constraints for skeletal point prediction.}
\label{tab:abl_skelpoints}
\end{table}

\begin{figure}[!t]
\vspace{-4mm}
    \centering      
  \begin{overpic}[width=0.95\linewidth]{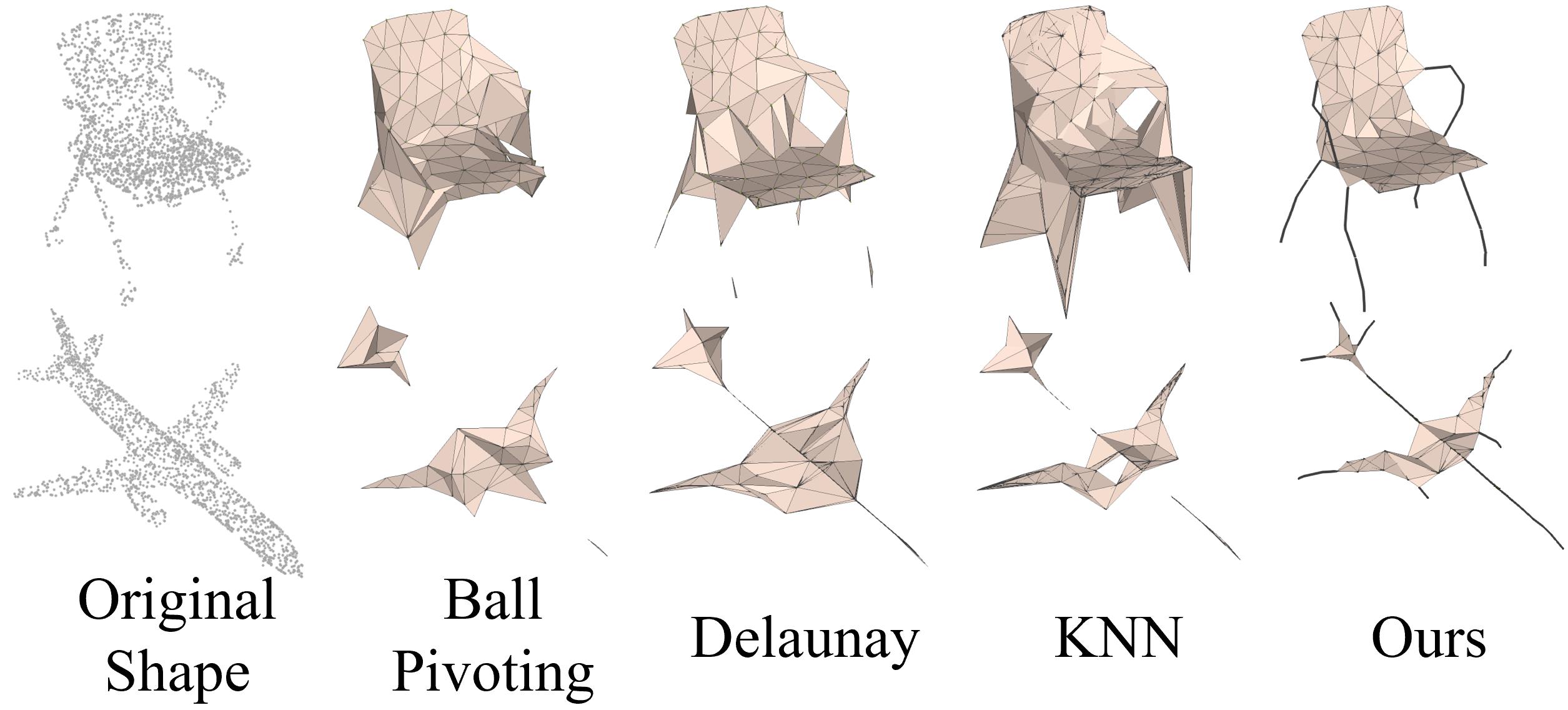}
    \end{overpic}
  \caption{Qualitative results of alternative mesh generation methods by connecting vertices to form mesh structures. }
  \label{fig:cmp_meshgen}
\end{figure}

\para{Mesh generation} Given a set of discrete points, there are some frequently used methods to generate meshes in the same manner as our method, i.e., connecting the vertices to form mesh structures. The alternative methods tested include Ball pivoting \cite{bernardini1999ballpivoting}, Delaunay triangulation (deleting the triangle faces with overlong edges using a threshold) and K Nearest Neighbor (KNN) (connecting the K nearest neighbors of each point and extracting the formed triangles). The qualitative and quantitative results are shown in Fig.~\ref{fig:cmp_meshgen} and Table~\ref{tab:abl_meshgen} (the first three rows). Our strategy for connecting points into a skeletal mesh is significantly better than the alternative methods, thanks to the joint use of the properties of skeletal mesh and the point correlations in latent embeddings; the other methods cannot preserve the typologies and thus are not suitable for generating skeletal meshes.

As an ablation study, in Table~\ref{tab:abl_meshgen}, we also report the errors without using the graph initialization (only skeletal spheres) and without using the GAE (only graph initialization), respectively. The results demonstrate the improvements brought about by the learned latent embeddings over the raw output of skeletal spheres and the initialization. 

\begin{table}[!htb]
\vspace{-2mm}
\resizebox{\columnwidth}{!}
{
\begin{tabular}{c|cccc}
\hline
              & CD-Recon        & HD-Recon        & CD-MAT          & HD-MAT          \\\hline
Delaunay      & 0.0388          & 0.3313          & 0.1196          & 0.3386          \\
Ball Pivoting & 0.0443          & 0.3903          & 0.1095          & 0.3965          \\
KNN           & 0.0394          & 0.2811          & 0.0946          & 0.3243          \\\hline
w/o graph     & 0.0525          & 0.1592          & 0.1060          & 0.2079          \\
w/o GAE       & 0.0383          & 0.1462          & 0.0842          & \textbf{0.1886} \\
w/ GAE        & \textbf{0.0372} & \textbf{0.1424} & \textbf{0.0828} & 0.1898                \\\hline
\end{tabular}
}
\caption{Quantitative comparisons with different methods for mesh generation by connecting vertices (first 3 rows); ablation study for the mesh generation (last 3 rows).}
\label{tab:abl_meshgen}
\vspace{-2mm}
\end{table}

\begin{figure}[!tb]
    \centering      
  \begin{overpic}[width=0.95\linewidth]{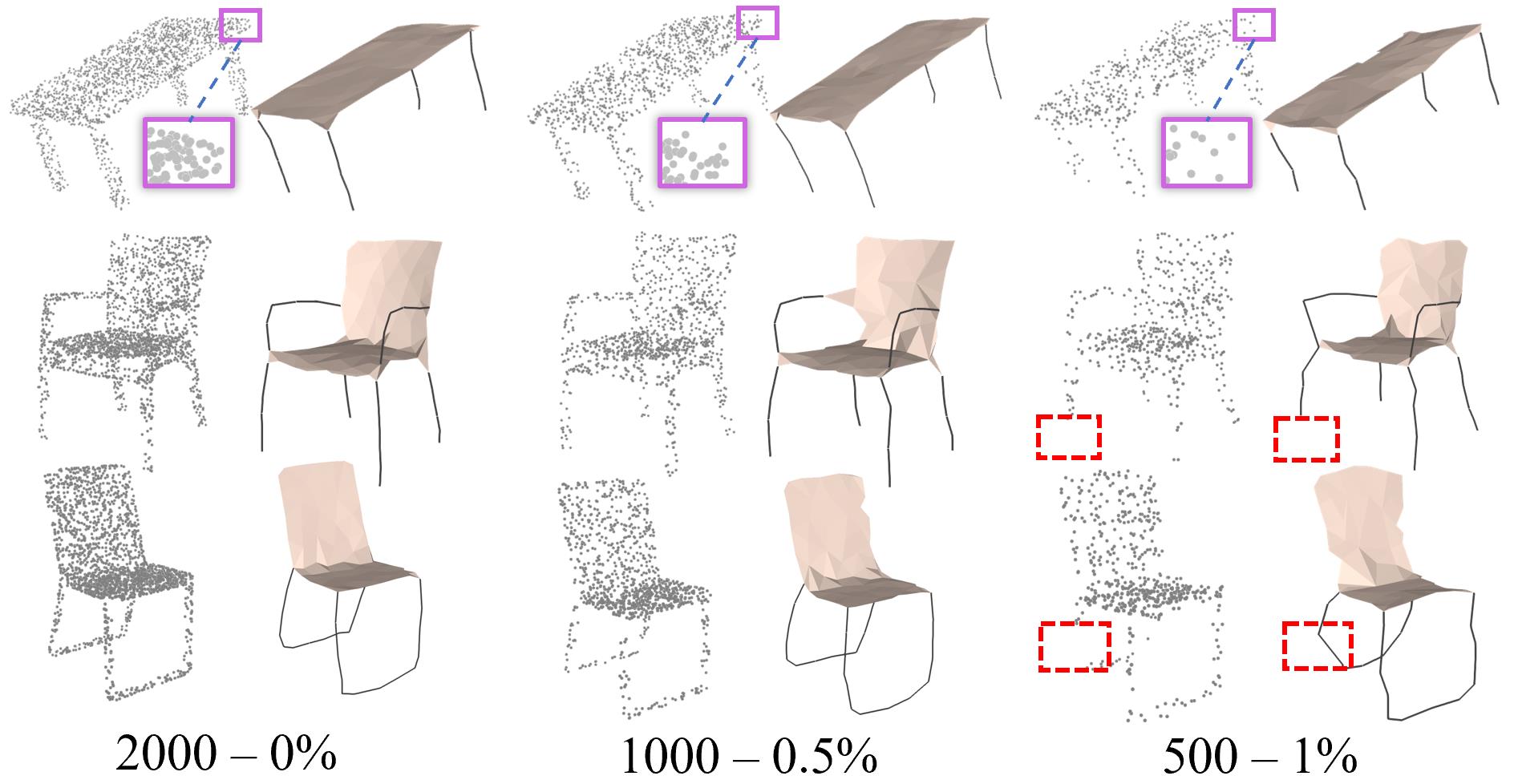}
    \end{overpic}
  \caption{Qualitative evaluation on the effect of the quality of input point cloud, i.e., density (point number) and noise.} 
  \label{fig:noise}
  \vspace{-6mm}
\end{figure}

\para{Noise and sparsity} To study the effect of the input quality on our algorithm, we evaluate our method using the input of different point numbers and noise levels, i.e., 2000 points with no noise, 1000 points with 0.5\% noise and 500 points with 1\% noise. The qualitative results are shown in Fig.~\ref{fig:noise} and more results can be found in the supplementary material, where we can find the results are largely similar.

However, since we use the convex combination of the input points to generate the skeletal points, if the input points are too sparse and some points on the convex hull are missing, our method struggles to recover the complete geometry (see the red boxes in Fig.~\ref{fig:noise}).
\begin{figure}[!tb]
\vspace{-2mm}
    \centering      
  \begin{overpic}[width=\linewidth]{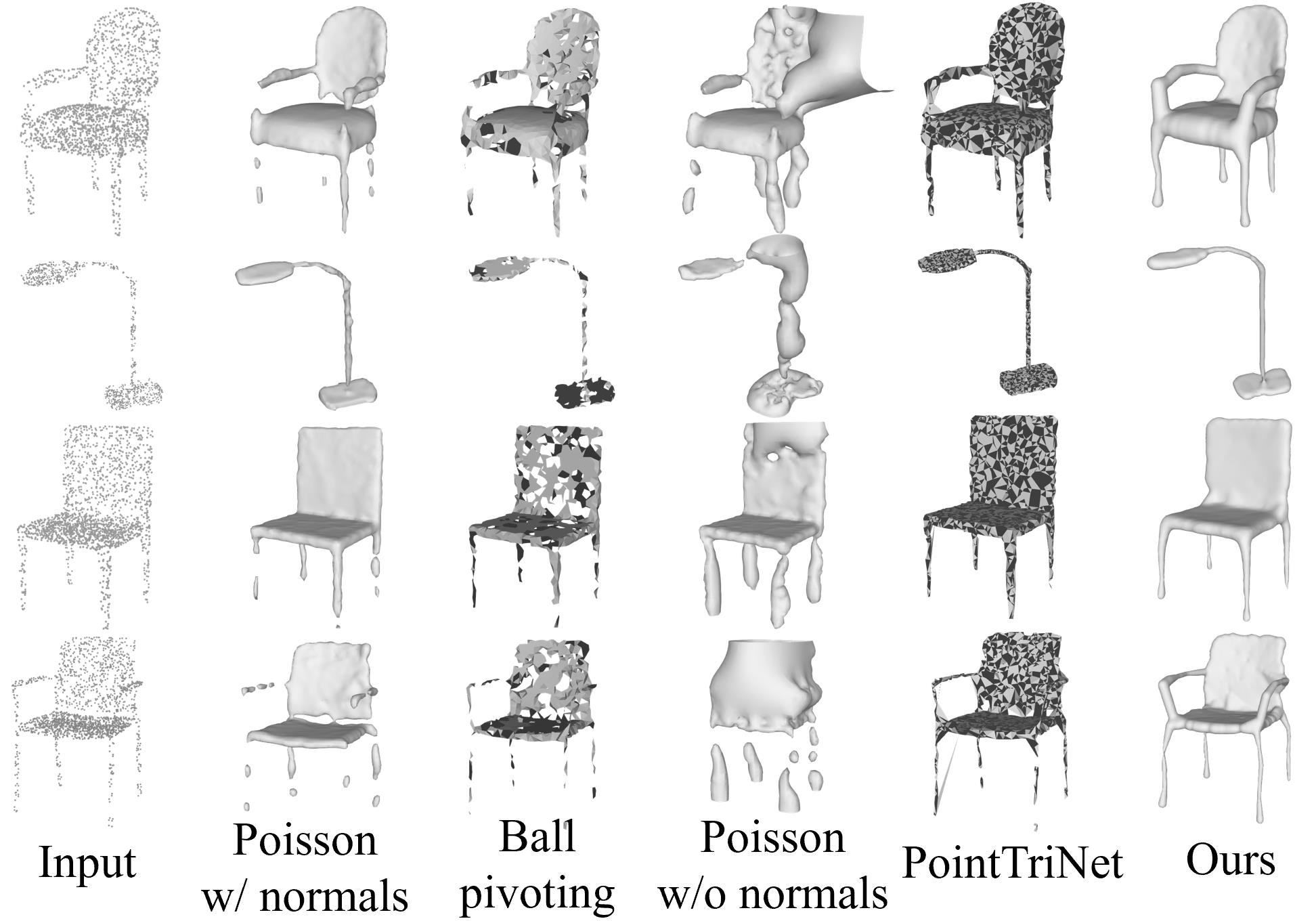}
    \end{overpic}
  \caption{Unsupervised surface reconstruction from point clouds. Our method can produce watertight surfaces without the need of surface normals and can capture the details of thin structures.}
\label{fig:app_recon}
\vspace{-4mm}
\end{figure}

\section{More Applications}
\label{sec:application}
Some good properties (see Sec.~\ref{sec:skl_mesh}) of the skeletal mesh, i.e., recoverability, abstraction, structure-awareness and volume-based closure, provide new insights into several unsupervised tasks on point clouds. In this section, we demonstrate the benefits to four such applications.

\para{Surface reconstruction} 
Generating watertight surfaces from point clouds for thin structures and preserving the original typologies, especially when there is no supervision from ground-truth, is a highly challenging problem. The skeletal mesh provides a suitable vehicle for solving this problem, since we can reconstruct the surface of an input point cloud by interpolating the skeletal spheres. We qualitatively compare to some unsupervised surface reconstruction methods, including Ball pivoting \cite{bernardini1999ballpivoting}, Screened Poisson reconstruction \cite{kazhdan2013screened} and  PointTriNet~\cite{sharp2020pointtrinet}. The results are shown in Fig.~\ref{fig:app_recon}. On the one hand, the reconstructions using skeletal meshes preserve the complex typologies and also capture the thin structures of the input. On the other hand, unlike the Poisson reconstruction, our method does not need to input any surface normal information, and is still able to produce high-quality watertight surfaces.

\begin{figure}[!tb]
    \centering      
    \vspace{-2mm}
  \begin{overpic}[width=1.0\linewidth]{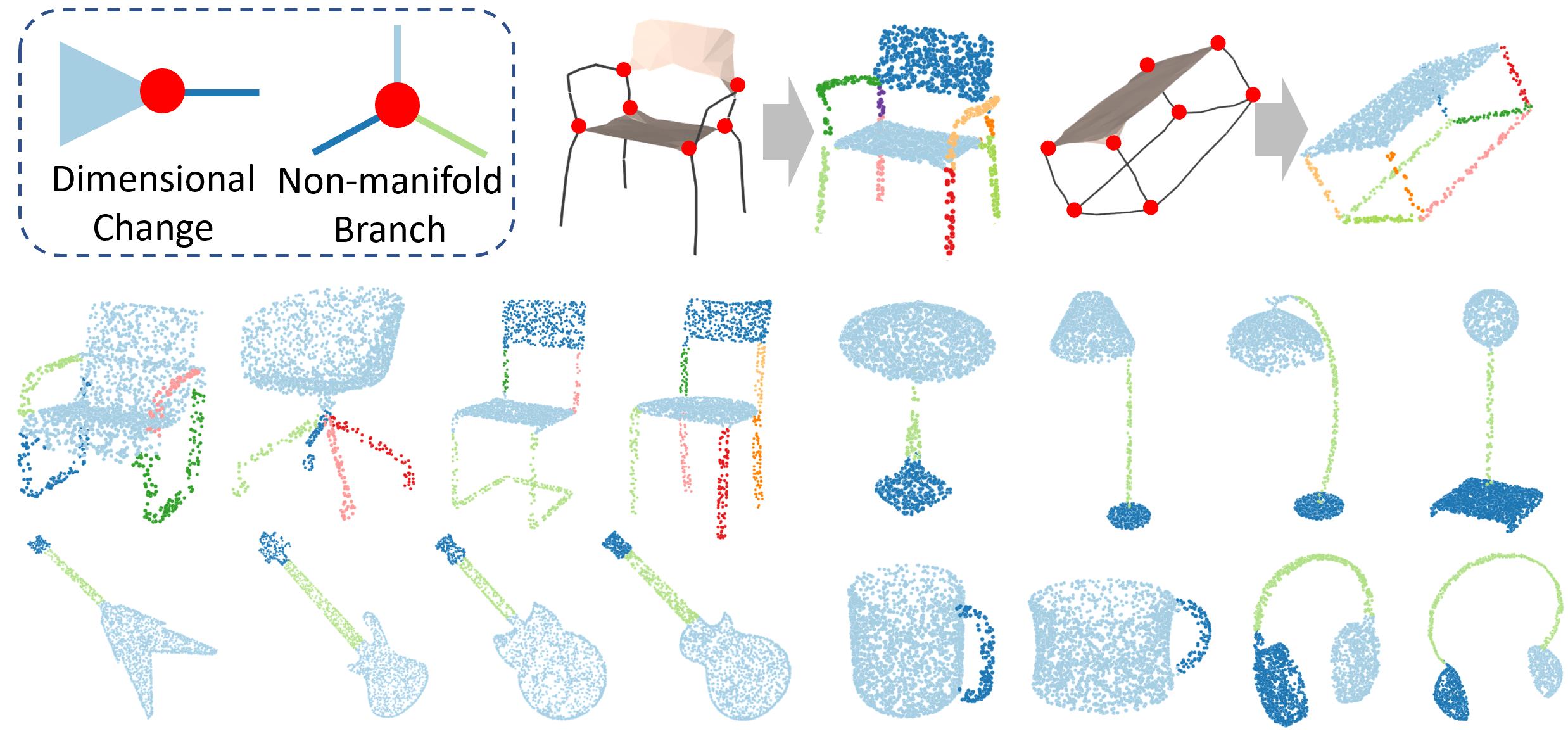}
    \end{overpic}
  \caption{Unsupervised structural decomposition for point clouds by detecting dimensional changes and non-manifold branches on the skeletal mesh.}
  \label{fig:seg}
\end{figure}

\para{Structural decomposition} The skeletal mesh is structure-aware as introduced in Sec.~\ref{sec:skl_mesh}; thus it naturally induces a structural decomposition of a shape without data annotation. Fig.~\ref{fig:seg} shows a set of segmentation results by detecting the dimensional changes and the the non-manifold branches on the skeletal meshes.

\begin{figure}[!tb]
    \vspace{-3mm}
    \centering      
  \begin{overpic}[width=0.98\linewidth]{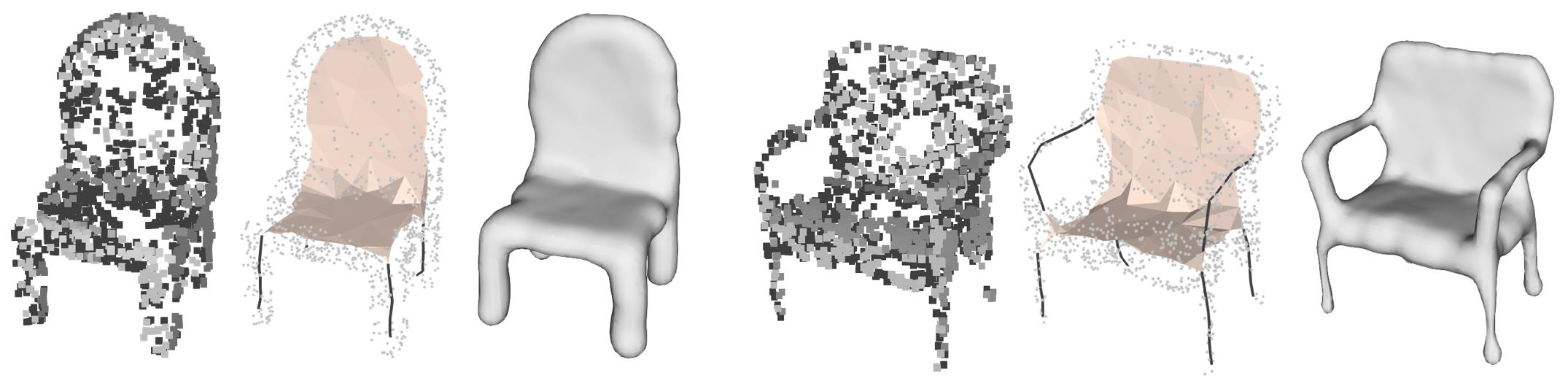}
    \end{overpic}
  \caption{Skeletal mesh prediction for reconstructing complete surfaces from point clouds with missing regions.}
  \label{fig:inpainting}
\end{figure}

\para{Shape in-painting} The skeletal mesh encodes volume-based shape context which provides better integrity of part geometry, leading to robustness against partially missing data. Thus we can recover the complete geometry from an incomplete point cloud. Some results are given in Fig.~\ref{fig:inpainting}.

\begin{figure}[!tb]
    \vspace{-3mm}
    \centering      
  \begin{overpic}[width=\linewidth]{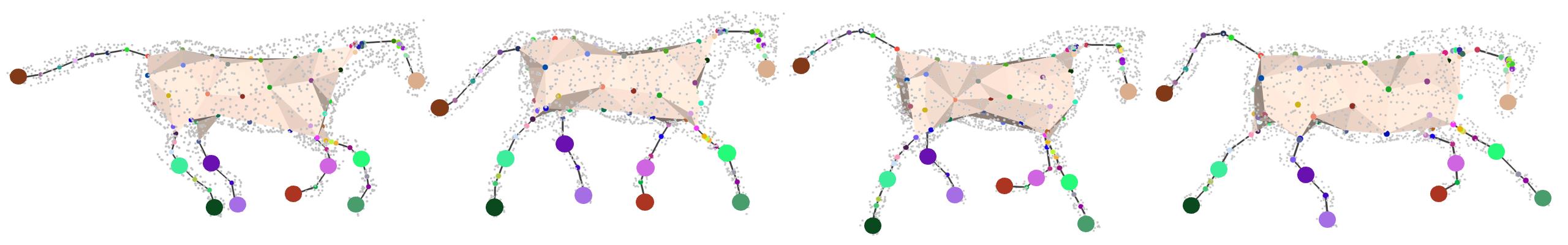}
   \end{overpic}
  \caption{Skeletal meshes with consistent correspondences. We color the skeletal points of each shape individually based on their indices in each prediction; some key skeletal points are highlighted to show the semantic consistency.}
    \label{fig:consistent_skel}
    \vspace{-5mm}
\end{figure}

\para{Skeleonization with consistent correspondence} Skeletons play a vital role in pose recognition and animation, in which a challenge is to find the correspondences between different poses of a human or an animal. Fig.~\ref{fig:consistent_skel} shows the skeletal meshes generated for a sequence of horse models in different poses. Without explicitly enforcing the correspondence, the results for different poses are semantically consistent due to the learning of convex combination (see \cite{chen2020unsupervised} for the detailed explanation). This would be beneficial for tasks that require pose invariance in animation. 

\section{Conclusion}
We propose {\em Point2Skeleton}, a novel unsupervised method for generating skeletal meshes from point clouds. We first predict skeletal points by learning a geometric transformation, and then analyze the connectivity of the skeletal points to form meshes. Experiments show that the skeletal mesh generated by our method effectively captures the underlying structures for general 3D shapes, even when represented as points clouds with noises or missing parts.

We believe our method for learning skeletons can benefit a variety of 3D applications, given the good properties of the skeletal mesh. In the future, how to combine the geometric and topological properties of the skeletal mesh and higher-level supervision from humans (e.g., semantics) for 3D learning tasks, would be a promising direction.

\para{Acknowledgement} We thank the anonymous reviewers for their valuable feedback. This work was supported by the General Research Fund (GRF) of the Research Grant Council of Hong Kong (17210718).

{\small
\bibliographystyle{ieee_fullname}
\bibliography{egbib}
}
\newpage

\begin{appendix}
\input{appendix_combine}
\end{appendix}
\end{document}

%% file: appendix_combine.tex
\section{Appendix}

\subsection{Rationale of Radius Computation}
Given a set of densely sampled points on the boundary surface of a 3D shape, since the shape must be inside its convex hull, it is easy to show that any arbitrary point inside the shape can be derived from a convex combination of the sampled points, i.e., a linear combination with non-negative weights $\{w_i\}$ summing up to 1.

\begin{figure}[!htb]
  \includegraphics[width=\linewidth]{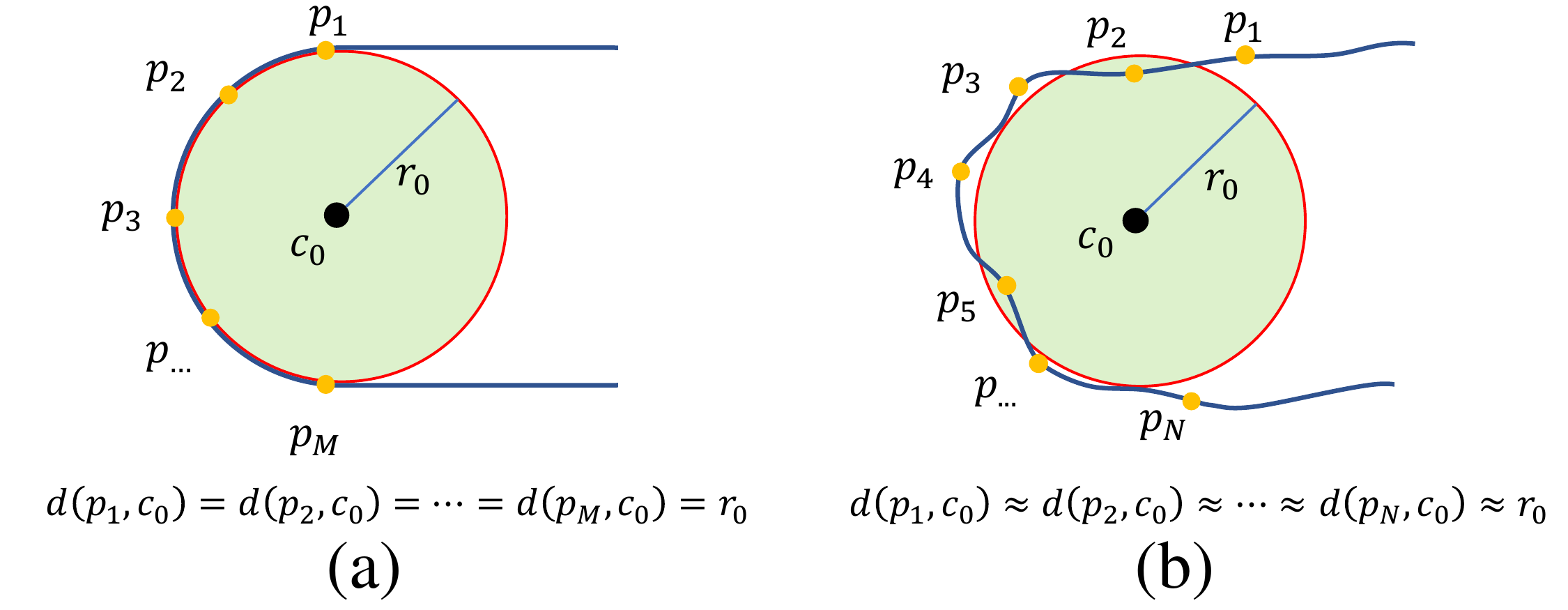}
  \caption{Illustration of the radius computation using closest distances. (a) An ideal case where $(c_0, r_0)$ is a maximal inscribed sphere; (b) a typical case where the surface has noise and the radius is approximated by the closest distances.}
  \label{fig:radius_compute}
\end{figure} 

Now we show why it is reasonable to use the same weights of convex combination to estimate the radius of a skeletal point based on their closest distances to the sampled points. Recall that, the closest distance from an input sample point $p$ to all the skeleton points $\{c_i\}$ is defined as,
\begin{equation}
    d(p, \{c_i\}) =  \min_{c \in \{c_i\}} \left\| p-c \right\|_2.
    \label{eq:r_dis}
\end{equation}

\begin{figure}[!t]
    \centering      
  \includegraphics[width=\linewidth]{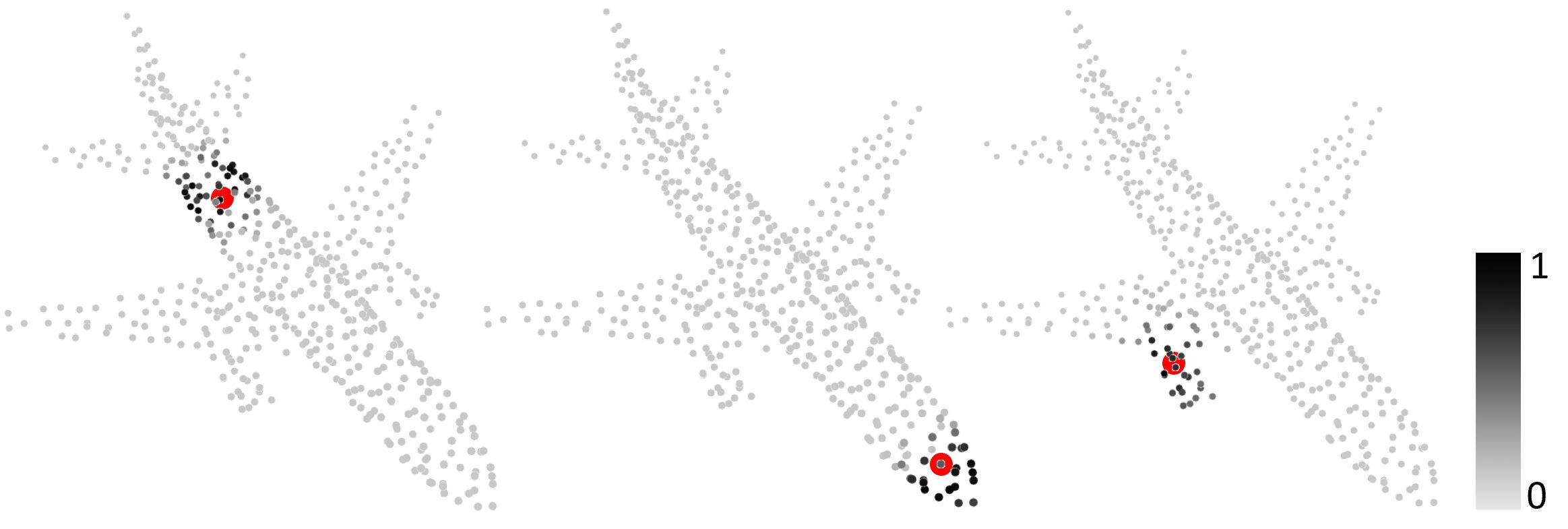}
  \caption{Visualization of the predicted combinational weights for the corresponding skeletal points. Each weight value $w_i$ is scaled by $w_i/w_{max}$ for better visualization, where $w_{max}$ is the maximum weight value for the corresponding skeletal point.}
  \label{fig:weights_vis}
\end{figure} 

To simplify the discussion, we first analyze an ideal case. As shown in Fig.~\ref{fig:radius_compute} (a), consider a skeletal sphere $s_0=(c_0,r_0)$ that is maximally inscribed in a shape, where $c_0$ is the coordinate of the sphere center and $r_0$ the sphere radius. Assume the sphere has $M$ ($M\ge 2$) touching points on the shape surface. Obviously, the $M$ points are the closest points to $c_0$ on the shape surface and their distances to the center $c_0$ are all equal to the sphere radius. Thus we have
\begin{equation}
    d(p_i, c_0) = r_0, \quad\quad \text{for}\quad  i=1,2,...,M.
\end{equation}

Then, given an arbitrary group of convex combination weights $\{w_1, w_2, ..., w_M\}$ of the $M$ points, if we use the weights to combine the closest distances, we always obtain a constant value which is the sphere radius:
\begin{equation}
   \sum_{i=1}^{M} {w_i}d(p_i, c_0)=r_0,  \quad \text{with}\quad \sum_{i=1}^{M}w_i=1.
\end{equation}

We now analyze the typical case to generate a skeletal point. First, we visualize the learned combinational weights of the input points for certain skeletal points in Fig.~\ref{fig:weights_vis}. For a given skeletal point $c$, although the combinational weights to derive $c$ are not unique, we observe that the weights are larger for the surface points that are very close to $c$, but are smaller or diminish to 0 for those far away from $c$. Therefore, the skeletal point $c$ is approximated by the convex combination of a local set of input points that are closest to $c$. As shown in Fig.~\ref{fig:radius_compute} (b), this is similar to the case discussed above. Hence, by using the same combinational weights, we approximate the radius by the weighted average of the closest distances, which therefore provides a reasonable estimation of the true radius at the skeletal point $c$.

\subsection{Network Architecture}
\begin{figure*}[t]
    \centering      
  \includegraphics[width=0.9\linewidth]{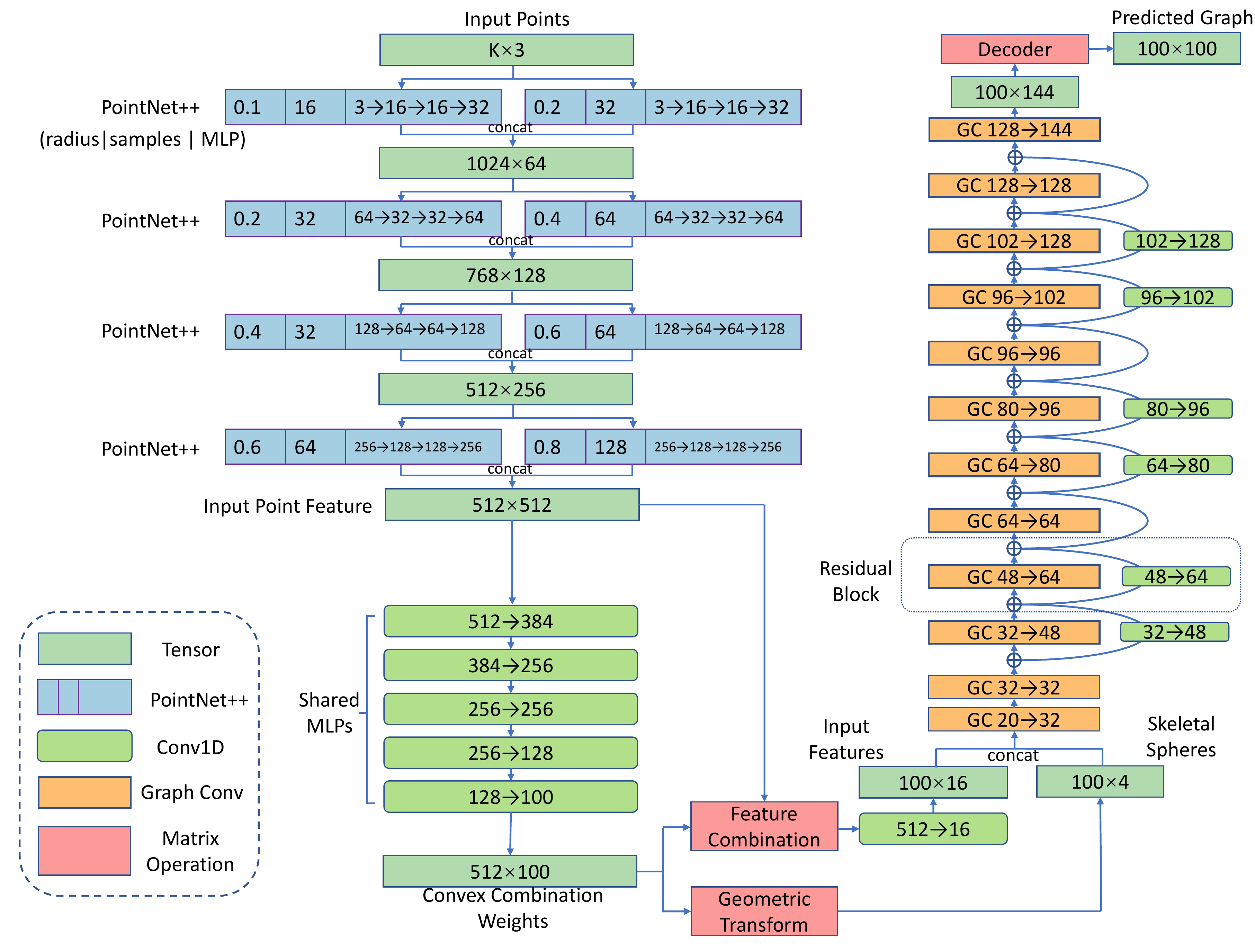}
  \caption{Overall network architecture of Point2Skeleton. }
  \label{fig:network}
\end{figure*}

Fig.~\ref{fig:network} shows the detailed network architecture of our method. We use PointNet++ \cite{qi2017pointnet++} as the encoder of the input point cloud, which is composed of 4 set abstraction (SA) levels. For each SA, we show the radius of the ball query, the number of local patches and the feature dimensions of the MLPs. We adopt a density adaptive strategy, i.e., multi-scale grouping (MSG), to combine the features from two different scales in each layer. The shared MLPs are for processing the contextual features encoded by the PointNet++ to predict the final combinational weights, where each layer is followed by a batch normalization and a ReLU non-linearity. We also use a dropout layer with a rate of 0.2 during training.

The contextual features of the input points are linearly combined using the predicted convex combination weights which serve as the input surface point features corresponding to a skeletal point. The combined contextual features are concatenated with the information of skeletal spheres (center coordinates and radii) to serve as the node features that are input to the GAE for link prediction. Each GCN layer is also followed by a batch normalization and a ReLU activation function. We use residual blocks \cite{he2016deep} between the consecutive GCN layers, where the additional branching layers are used to align the dimension of features.

\subsection{Graph AutoEncoder}

Given an undirected and unweighted graph with $N$ nodes, the encoder is defined as a series of graph convolutional layers. Unlike most existing works that use a shallow GCN (usually no more than 4 layers \cite{wu2020comprehensive}), we use a deep GCN with 12 layers to capture richer structures at various levels of abstraction. Consequently, to handle the degeneration problem caused by the depth of the network, we also include residual blocks \cite{he2016deep} between consecutive layers. The encoder is given by:
\begin{equation}
 \begin{aligned}
       & \quad \quad \quad GCN(\bm{X}^0,\bm{A})={\tilde{\bm{A}}}\bm{X}^{L-1}{\bm{W}}^{L-1}, \text{with\;} \\
         & \bm{X}^l= \sigma({\tilde{\bm{A}}}\bm{X}^{l-1}{\bm{W}}^{l-1} + \bm{X}^{l-1})  \text{, for\;} l\in\{1,...,L-1\}.
\end{aligned}
\end{equation}

Here, $\bm{A}\in \{0, 1\}^{N\times N}$ is the adjacency matrix, and $\tilde{\bm{A}}$ is the symmetrically normalized $\bm{A}$ given by $\tilde{\bm{A}}=\bm{D}^{-\frac{1}{2}}(\bm{A}+\bm{I}_N)\bm{D}^{-\frac{1}{2}}$, where $\bm{D}$ is the degree matrix of $\bm{A}$ and $I_N$ the identity matrix indicating self-connections. $\bm{X}^0$ is the input node features and $\bm{X}^l$ the latent features. $\bm{W}^l$ is a layer-specific trainable weight matrix. $\sigma$ is the ReLU activation function and $L$ is the number of layers. 
The decoder we use is a simple inner product decoder to produce the reconstructed adjacency matrix $\hat{\bm{A}}$:
\begin{equation}
    \hat{\bm{A}}=Sigmoid(\bm{Z}\bm{Z}^T) \quad \text{with} \quad {\bm{Z}}=GCN(\bm{X}^0,{\bm{A}}),
\end{equation}
where $\bm{Z}$ is the learned latent features. By applying the inner product on the latent variables $\bm{Z}$ and $\bm{Z}^T$, we measure the similarity of each node inside $\bm{Z}$. The larger the inner product $z_i^Tz_j$ in the embedding is, the stronger correlation the nodes $i$ and $j$ exhibit, which indicates that they are more likely to be connected.

\begin{figure}[!htb]
\centering
  \includegraphics[width=0.9\linewidth]{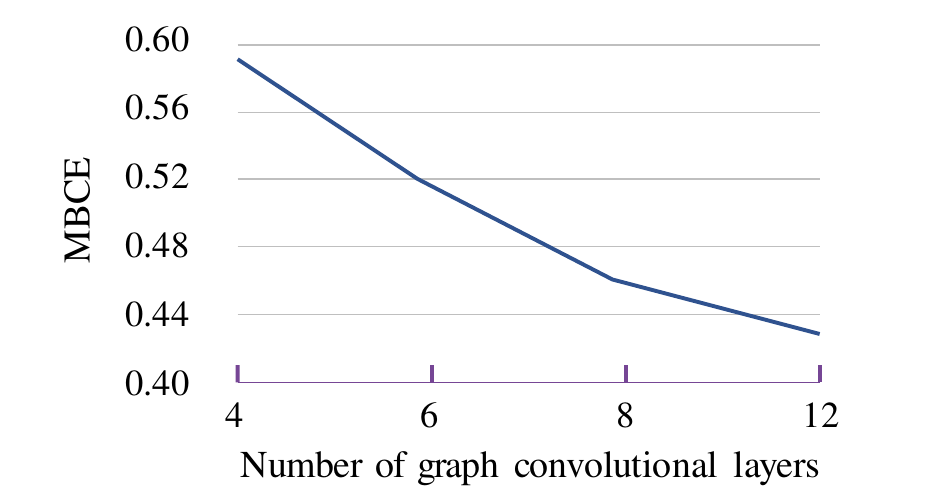}
  \caption{Evaluation of the number of graph convolutional layers. }
  \label{fig:gcn_layers}
\end{figure}

We evaluate the effect of the number of the graph convolution layers. As shown in Fig.~\ref{fig:gcn_layers}, a deeper GCN achieves better performance, i.e., smaller Masked Balanced Cross-Entropy (MBCE) loss \cite{tran2018pred_ae}.

\begin{figure*}[!t]
    \centering      
  \includegraphics[width=0.85\linewidth]{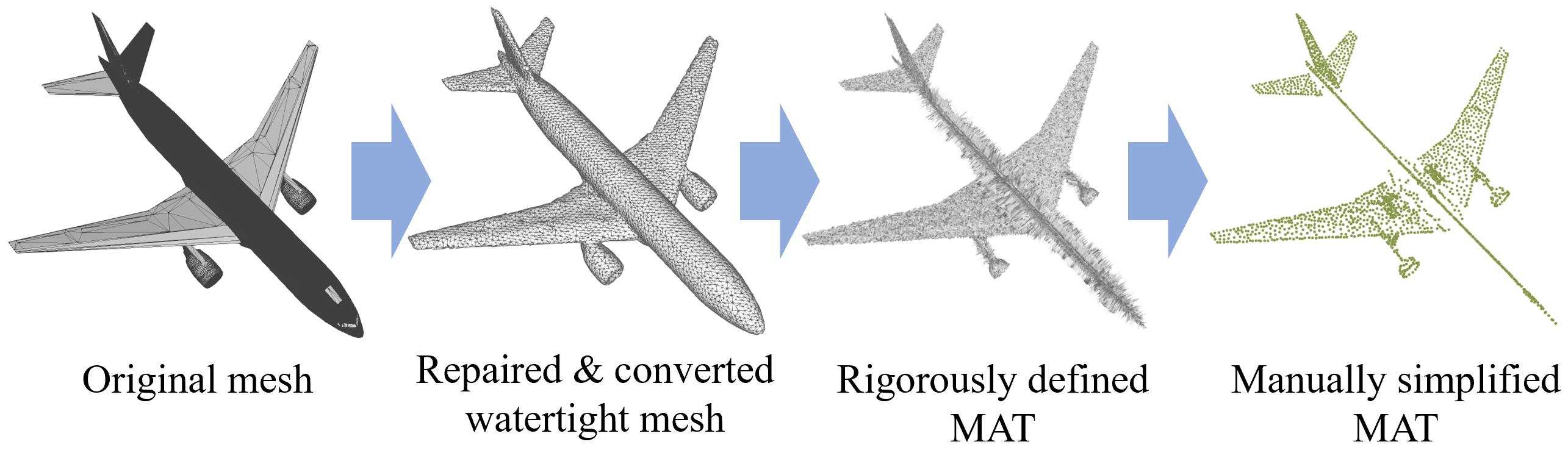}
  \caption{Steps to compute a simplified MAT from a surface mesh.}
  \label{fig:MAT_simplify}
\end{figure*} 

\subsection{Standard MAT for Evaluation}

\para{Computation.} As mentioned in the paper, there are no existing metrics to evaluate whether a skeletonization is reasonable. We use the manually simplified MAT that not only has meaningful structures but also exhibits good geometric accuracy, to evaluate different methods.

\begin{figure}[!htb]
  \includegraphics[width=\linewidth]{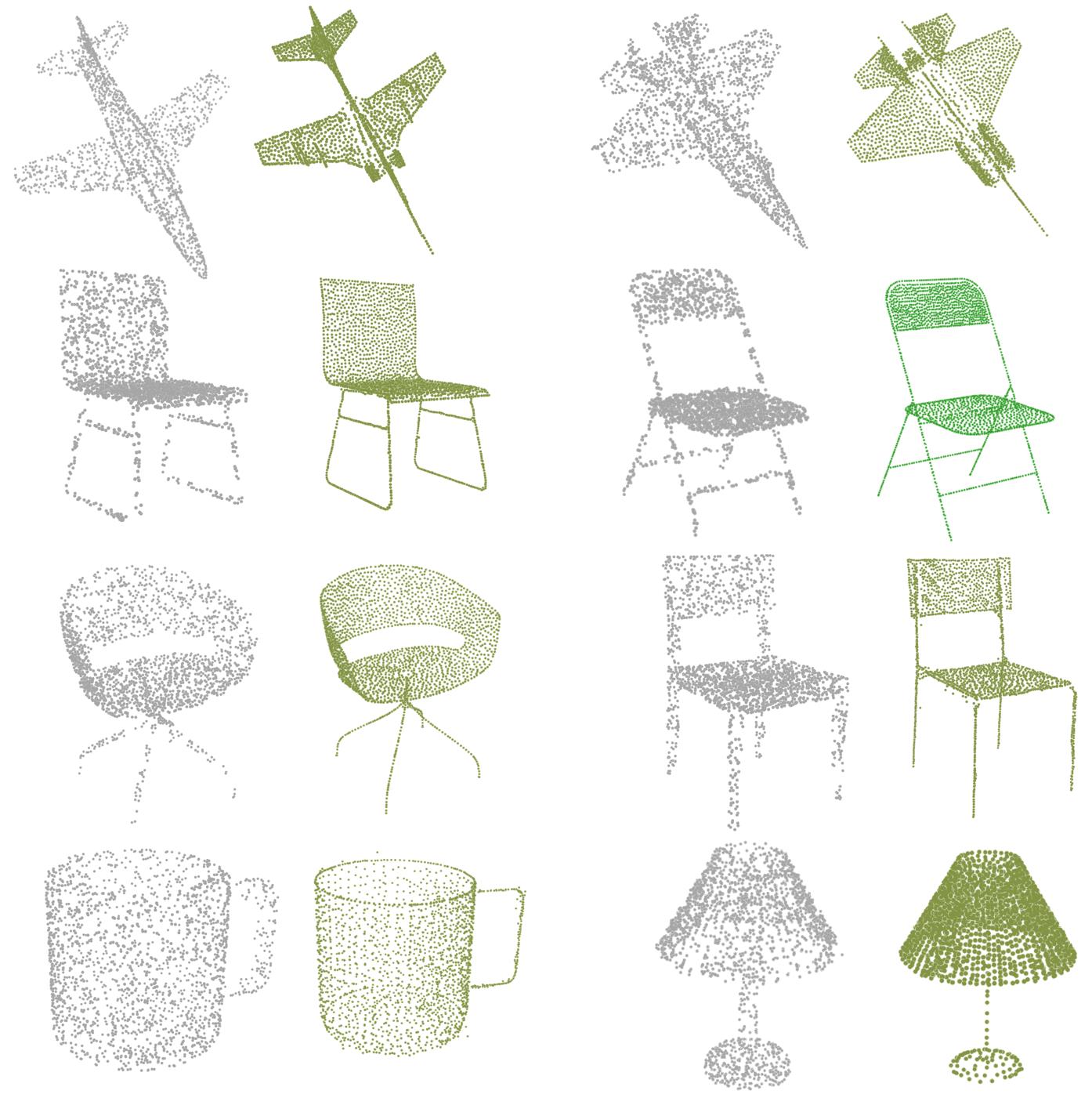}
  \caption{More examples of the simplified MAT used for evaluation.}
  \label{fig:mat_more_eg}
\end{figure} 

The main steps we adopt to generate a simplified MAT are shown in Fig.~\ref{fig:MAT_simplify}. For a point cloud in our dataset, we first find its ground-truth mesh in the ShapeNet \cite{chang2015shapenet}. Since the original mesh in the ShapeNet is not watertight either, we need to repair and convert the mesh to a strictly watertight one \cite{huang2018manifold}. Then, we compute a standard MAT, which inevitably contains numerous insignificant spikes. After that, we manually simplify the MAT using a rule-based method \cite{li2015qmat}; we choose a certain threshold by which most spikes can be removed and the result is clean and structurally meaningful. Finally, we sample points on the simplified MAT for evaluating the CD and HD. 

Fig.~\ref{fig:mat_more_eg} shows more examples of the simplified MAT used for evaluation. It can be observed that the simplified MATs precisely capture the underlying structures of the input shapes using curve-like and surface-like components. Thus, they are suitable for evaluating whether the skeletonization of a method is reasonable and accurate.

These experiments, on the other hand, show that computing a clean and structurally meaningful skeletal representation by MAT simplification is difficult and expensive, given these tedious and time-consuming steps of geometric processing. By comparison, our method is more feasible, easier to use, and more efficient in practical applications.

\para{Can we compute the MAT from the surface mesh reconstructed from a point cloud?} Note that the simplified MATs in our dataset are computed by the ground-truth meshes corresponding to the point clouds. Considering that the surface mesh can also be reconstructed from a point cloud, one may wonder if we can directly reconstruct the surface meshes from the point clouds, and then compute the MATs based on the reconstructed surfaces. Through the experiments, we find this strategy is not feasible, since the surface quality of the existing surface reconstruction methods (like Poisson reconstruction) cannot satisfy the requirement of MAT computation. As a result, the MAT computation algorithm crashes in most cases.

\subsection{Effect of Input Quality}
\begin{figure}[!htb]
  \includegraphics[width=\linewidth]{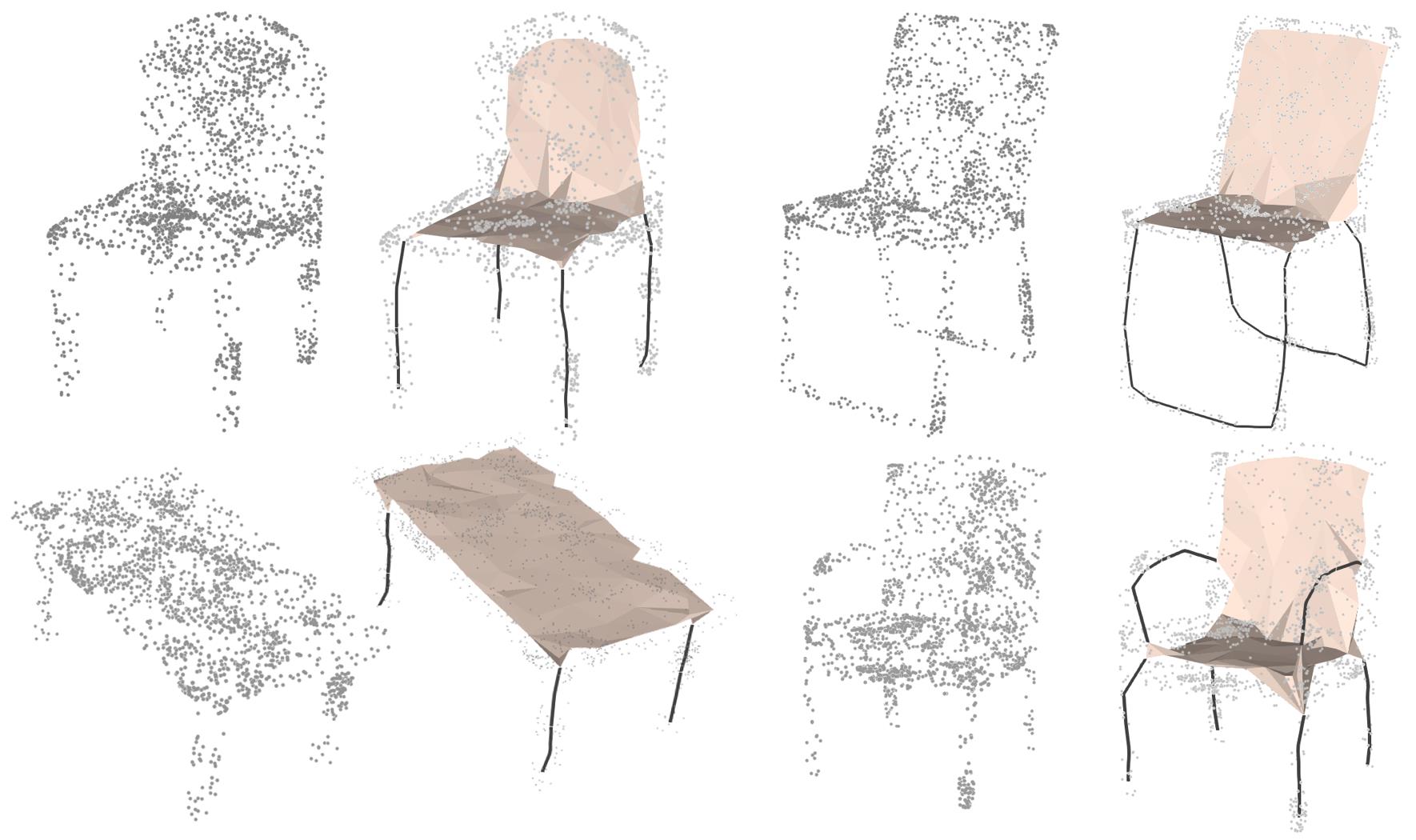}
  \caption{Generated skeletal meshes for point clouds with uneven density distribution.}
  \label{fig:uneven}
\end{figure}

We show more detailed quantitative results for evaluating the effect of the input quality to our method. As shown in Table~\ref{tab:noise}, we input point clouds with different point numbers and noise levels, i.e., 2000 points without noise, 1000 points with 0.5\% noise, and 500 points with 1\% noise, to our method; the quantitative results are largely similar. 
Besides, we test some point clouds with uneven density distribution, of which results are given in Fig.~\ref{fig:uneven}; our method is robust to the density variation of point cloud. 

\begin{table}[!htb]
\resizebox{\columnwidth}{!}
{\begin{tabular}{c|cccc}\hline
           & \multicolumn{1}{c}{CD-Recon} & \multicolumn{1}{c}{HD-Recon} & \multicolumn{1}{c}{CD-MAT} & \multicolumn{1}{c}{HD-MAT} \\\hline
2000 - 0\%   & 0.0372   & 0.1424   & 0.0828 & 0.1898 \\
1000 - 0.5\% & 0.0382   & 0.1615   & 0.0851 & 0.2071 \\
500 - 1\%    & 0.0458   & 0.2127   & 0.0958 & 0.2524   \\\hline
\end{tabular}
}
\caption{Quantitative evaluation results on different number of input points and different levels of noise.}
\label{tab:noise}
\end{table}

\begin{figure}[!htb]
    \centering      
  \begin{overpic}[width=\linewidth]{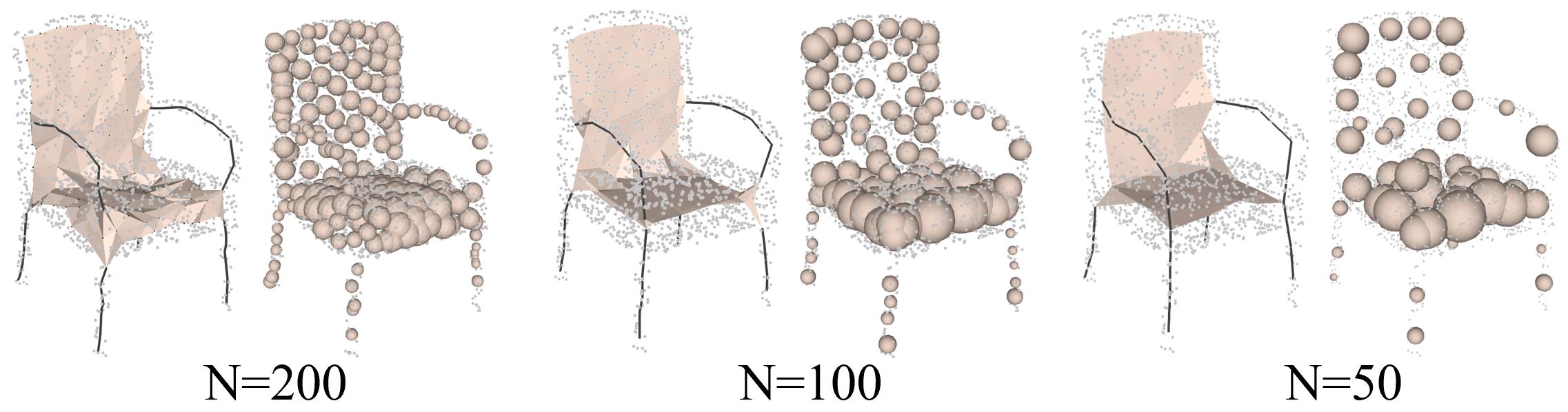}
    \end{overpic}
\caption{Results of skeletal mesh prediction using different number of skeletal points ($N$).} 
  \label{fig:N}
\end{figure} 

\subsection{Effect of Skeletal Point Number}
To study how the number of skeletal points ($N$) affects the results of skeletal mesh generation, we use $N=200, 100$ and $50$ for evaluation. The qualitative and quantitative results are given in Fig.~\ref{fig:N} and Table~\ref{tab:N}. The results suggest that, with the other conditions unchanged, using more skeletal points leads to lower reconstruction error but includes more insignificant details, which reduces the simplicity and abstraction level. To balance the accuracy and structural simplicity, we use $N=100$ in the paper.

\begin{table}[!htb]
\resizebox{\columnwidth}{!}
{\begin{tabular}{c|cccc}\hline
           & \multicolumn{1}{c}{CD-Recon} & \multicolumn{1}{c}{HD-Recon} & \multicolumn{1}{c}{CD-MAT} & \multicolumn{1}{c}{HD-MAT} \\\hline
$N$ = 200 & 0.0310   & 0.1402   & 0.0782 & 0.2002 \\
$N$ = 100 & 0.0372   & 0.1424   & 0.0828 & 0.1898 \\
$N$ = 50  & 0.0494   & 0.1720   & 0.0990 & 0.2225\\\hline
\end{tabular}
}
\caption{Quantitative evaluation results on different number of skeletal points ($N$).}
\label{tab:N}
\end{table}

\subsection{Limitations and Failure Cases}
For the skeletal point prediction, as mentioned in the paper, we use convex combination of the input points to generate the skeletal points. Therefore, our method will fail to recover a partial point cloud if its original skeleton cannot be completely included in the convex hull of the partial shape.

\begin{figure}[!htb]
  \includegraphics[width=\linewidth]{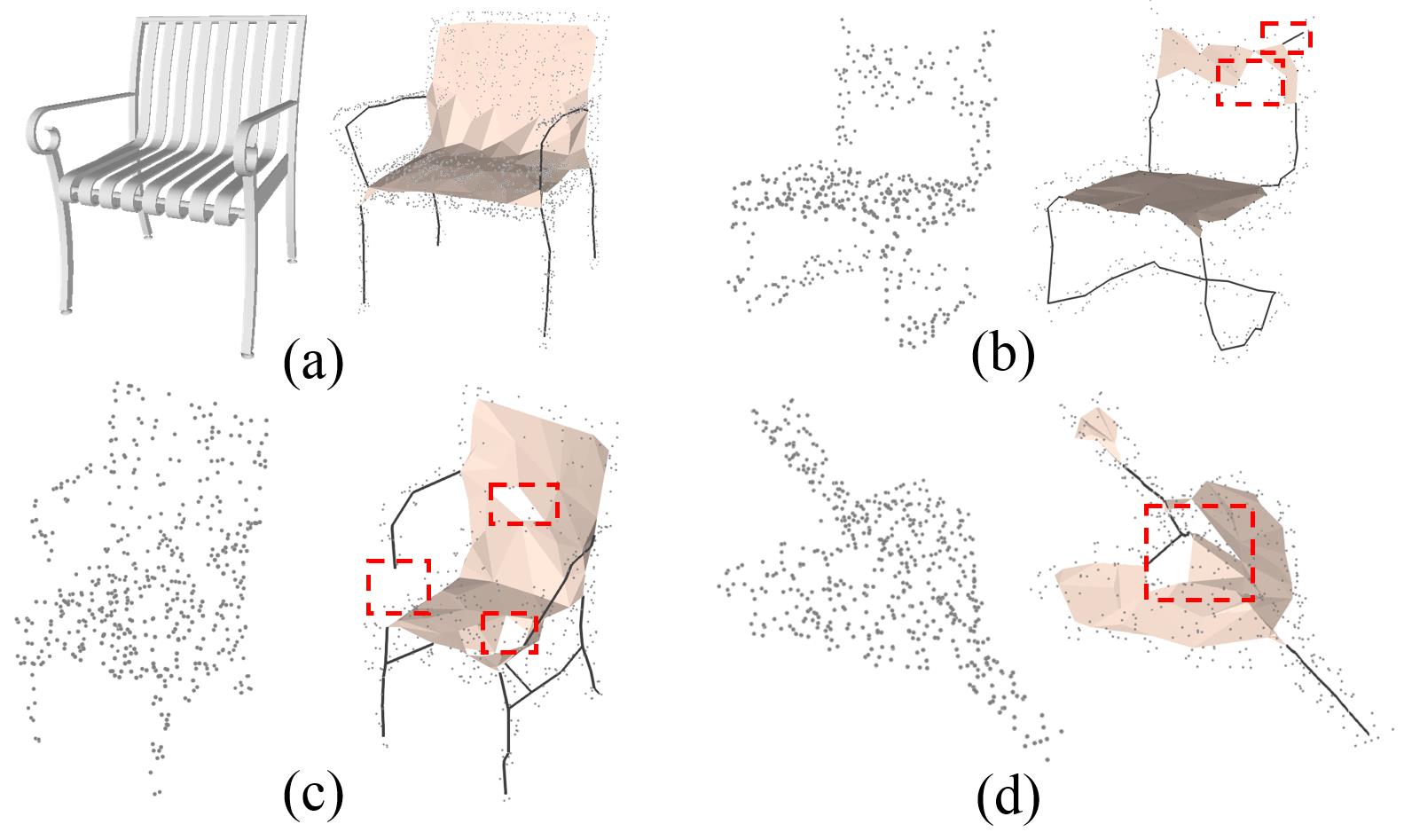}
  \caption{Failure cases of the skeletal mesh generation. }
  \label{fig:limitation}
\end{figure} 
Generally, we observe the prediction of skeletal points is stable driven by the shape geometry, while connecting the skeletal points to faithfully capture the shape structure is more challenging. We show some failure cases of the mesh generation in Fig.~\ref{fig:limitation}. First, we use local connectivity as a prior to initialize the edge connections of the graph. However, the local connectivity is not always true but based on the assumption that the point cloud has a relatively distinguishable structure. As shown in Fig.~\ref{fig:limitation}~(a), solely using local connectivity cannot enforce all the connections to be located inside the shape, leading to inconsistent structures with the original shape. Giving additional inside/outside labels as supervision like \cite{chen2019learningimplicit,genova2019shapetemplate} would be a potential solution. 

Second, although we use some strategies to make the mesh generation more reliable, some results still have unsmooth or incorrect structures, especially when the point cloud is noisy and sparse, as shown in Fig.~\ref{fig:limitation}~(b)(c)(d). On the one hand, without ground truth data, this is an inherent challenge for the unsupervised learning to exactly capture the detailed shape geometry. On the other hand, the link prediction of GAE is only based on the correlations of the skeletal points in the latent space, of which learned features remain not fully explained. Incorporating explicit topological constraints in the network would help improve the mesh quality.